\documentclass[journal,twoside,web]{ieeecolor}
\usepackage{jsen}
\usepackage{amsmath,amssymb,amsfonts}
\usepackage{graphicx, caption}
\usepackage{textcomp}
\usepackage{wrapfig}
\usepackage{algorithm}
\usepackage[noend]{algpseudocode}
\usepackage[style=numeric, backend=biber, sorting=none]{biblatex}
\usepackage[export]{adjustbox}
\usepackage{dblfloatfix}
\usepackage{gensymb}
\usepackage{subcaption}
\usepackage{stmaryrd}
\usepackage{varwidth}
\usepackage{blindtext}
\usepackage{scrextend}
\usepackage{xurl}
\usepackage{comment}
\usepackage{placeins}
\usepackage{multirow}
\usepackage[normalem]{ulem}
\useunder{\uline}{\ul}{}

\newcommand\floor[1]{\lfloor#1\rfloor}

\def\BibTeX{{\rm B\kern-.05em{\sc i\kern-.025em b}\kern-.08em
    T\kern-.1667em\lower.7ex\hbox{E}\kern-.125emX}}
\definecolor{abstractbg}{rgb}{0.89804,0.94510,0.83137}
\begin{document}
\title{SSCATeR: Sparse Scatter-Based Convolution Algorithm with Temporal Data Recycling for Real-Time 3D Object Detection in LiDAR Point Clouds}
\author{Alexander Dow, Manduhu Manduhu, Matheus Santos, Ben Bartlett, Gerard Dooly and James Riordan \ 
\thanks{© 2026 IEEE.  Personal use of this material is permitted.  Permission from IEEE must be obtained for all other uses, in any current or future media, including reprinting/republishing this material for advertising or promotional purposes, creating new collective works, for resale or redistribution to servers or lists, or reuse of any copyrighted component of this work in other works.}
\thanks{This work was supported by the RAPID project, funded through the European Commission’s Horizon 2020 research and innovation programme under Grant Agreement number 861211.}
\thanks{Alexander Dow, Manduhu Manduhu and James Riordan are with the Drone Systems Lab, School of Computing, Engineering and Physical Sciences, University of the West of Scotland, Blantyre, Glasgow, G72 0LH, UK (e-mail: alex.dow@uws.ac.uk, manduhu.manduhu@uws.ac.uk, james.riordan@uws.ac.uk). }
\thanks{Matheus Santos, Ben Bartlett and Gerard Dooly are with the Centre for Robotics and Intelligent Systems, University of Limerick, Limerick, V94 T9PX, Ireland.}}

\IEEEtitleabstractindextext{%
\fcolorbox{abstractbg}{abstractbg}{%
\begin{minipage}{\textwidth}%
\begin{wrapfigure}[13]{r}{3.5in}%
\includegraphics[width=3.5in, height=1.5in]{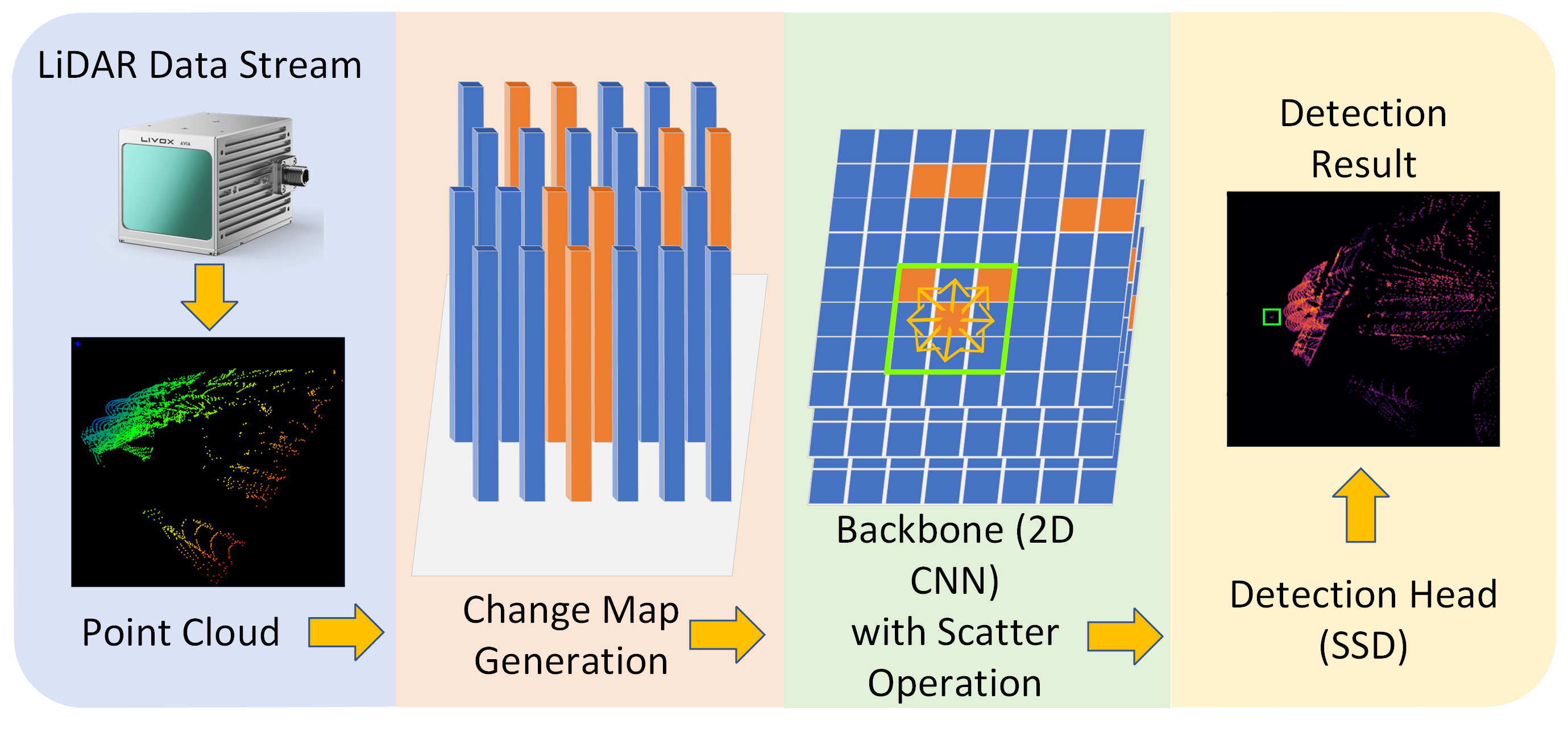}%
\end{wrapfigure}%

\begin{abstract}

This work leverages the continuous sweeping motion of LiDAR scanning to concentrate object detection efforts on specific regions that receive a change in point data from one frame to another. We achieve this by using a sliding time window with short strides and consider the temporal dimension by storing convolution results between passes. This allows us to ignore unchanged regions, significantly reducing the number of convolution operations per forward pass without sacrificing accuracy. This data reuse scheme introduces extreme sparsity to detection data. To exploit this sparsity, we extend our previous work on scatter-based convolutions to allow for data reuse, and as such propose Sparse Scatter-Based Convolution Algorithm with Temporal Data Recycling (SSCATeR). This operation treats incoming LiDAR data as a continuous stream and acts only on the changing parts of the point cloud. By doing so, we achieve the same results with as much as a 6.61-fold reduction in processing time. Our test results show that the feature maps output by our method are identical to those produced by traditional sparse convolution techniques, whilst greatly increasing the computational efficiency of the network.

\end{abstract}

\begin{IEEEkeywords}
crosstalk interference, object detection, point cloud, PointPillars, scatter operation, sparse convolution, streaming LiDAR.
\end{IEEEkeywords}
\end{minipage}}}

\maketitle
\section{Introduction}
\label{sec:introduction}
\IEEEPARstart{S}{ense} and Detect (SAD) is crucial for autonomous drones and other autonomous vehicles in dynamic environments \cite{sense_and_avoid}. Sense is defined as a robot's ability to monitor its surroundings and detect is defined as its ability to identify and analyze the risk posed by potential threats and obstacles. Minimizing SAD latency enhances safety, as it enables quicker recognition and response to hazards.

As such, SAD is especially critical when immediate actions are required to avoid collisions between drones operating in a swarm. A popular use case is maintenance inspection of large civil engineering infrastructure, where multi-drone swarms can greatly improve survey efficiency by performing multiple data collection missions simultaneously \cite{lidarsim}. Although their flight plans are strategically separated using trajectory-based planning, the drones can be required to fly paths that risk denial of global navigation satellite system (GNSS) and communications due to multi-path and signal occlusion. The likely deterioration of these external systems requires drone-embedded solutions to mitigate the collision risk associated with two drones flying in close proximity and potentially with poor positional estimation. 

Drone-embedded light detecting and ranging (LiDAR) sensors are well suited for SAD tasks conducted in situations such as the above use case \cite{manduhu2023airborne}. The range of drone-embedded LiDAR for object detection is limited, often to less than 100 m, but this is still suitable for swarms conducting maintenance inspections, where they will be operating with small separation distances from each other and the infrastructure itself. The sensor's time-of-flight mode of operation provides native depth information needed for three dimensional (3D) localization and object range estimation and proves robust in low light conditions. Nonetheless, addressing the high latencies involved in constructing 3D point clouds and performing point-based object detection remains a significant challenge.

Although LiDAR sensors continuously collect and process data points, the resultant point clouds take time to accumulate enough data to make object detection viable, which introduces latency. As such, the collection of discrete laser channel updates that occur within a specified time window are integrated into 3D point cloud "frames". The high LiDAR pulse repetition rate ensures substantial frame-to-frame coherency between scan cycle frames. By using a sliding time window, with each advancement of the time window, the newest points are incorporated into the "frame," displacing the oldest points. Aggregating multiple scan cycles over extended time windows yields increasingly detailed point clouds with dense focal points. Significantly, in successive frames much of the global point cloud remains unchanged.

Whilst increasing point density using longer time windows can have a positive impact on accuracy in point-based object detection, it also adds further latency through increased processing time. In point-based object detection, Convolutional Neural Network (CNN) models are commonly employed to handle structured grid-like data, such as "pseudoimages". Pillar-based and voxel-based feature extraction methods bridge the gap between pixel and point modalities, allowing CNN-based models to process unstructured point cloud data effectively. Sparse convolution techniques are commonly used to handle the sparsity of 3D point cloud data, improving computational efficiency \cite{Submanifold}. However, while sparse convolution techniques exploit spatial sparsity, they overlook the temporal aspect of LiDAR scanning patterns. With much of the cloud being unchanged, even those point-based object detection networks which employ sparse convolution are therefore repeating many redundant operations in successive forward passes through the network.

Our work synchronizes the convolution calculations in object detection with the LiDAR scanning pattern. We advance a sliding time window ($100\ ms$) of point data by short strides ($10\ ms$) and reuse prior convolution results in unchanged regions of the point cloud. This concentrates the computation efforts on the most recent changes in point data and significantly reduces the number of convolution operations per forward pass without sacrificing accuracy. We call the algorithm "Sparse Scatter-Based Convolutions with Temporal Data Recycling" (SSCATeR).

The contributions of this work are as follows:
\begin{enumerate}
  \item In our drone object detection use case we reduce the processing time for the convolutional backbone of Aerial-PointPillars by an average of $59.85\%$, with as much as an $84.88\%$ decrease over a single convolution layer. This reduction is due to the $72.80\%$ average reduction in active sites that require processing per frame by using the point timestamps to automatically reuse convolution calculations in unchanged regions. 
  \item Our results confirm that the accelerated convolution algorithm does not modify the outputs of each convolution layer, and thus the accuracy of the model is unaffected. We incorporated the algorithm into the PointPillars \cite{pointpillars} deep neural network by modifying its feature extraction and backbone stages, building on our previous work \cite{manduhu2023airborne} that embedded the model on a drone swarm to mitigate airborne collision risk. The end-to-end model achieves an mAP of $89.738\%$ and a recall of $91.277\%$.
\end{enumerate}

\section{Related Work}
\label{sec:related work}

\subsection{Point-Based Object Detection}
Popular point-based object detection networks such as PointPillars~\cite{pointpillars} and the significantly slower VoxelNet~\cite{voxelnet} discretize input point cloud frames into pillars and voxels respectively for feature extraction. They handle the input frame data as generic point clouds. The data structuring process is repeated for each forward pass, with the previous discretizations discarded.

PointPillars is a network known for both speed and accuracy, and has been the subject of extensive modification attempts to improve these two metrics. However, as this section will show, the trade off between the two has meant that research has consistently improved one at the detriment to the other, or to the detriment of safety critical features. PointRCNN~\cite{pointrcnn}, for example, has its inference speed improved by Fast Point-RCNN~\cite{fast_point_rcnn} from 10 FPS to 15 FPS, but when compared on the KITTI dataset, PointRCNN records a higher mAP score. When it comes to resource restricted embedded systems, there is a need to increase the speed of fast and accurate networks such as PointPillars without decreasing that accuracy.

McCrae and Zakhor~\cite{mccrae_3d_2020} presented an approach to PointPillars using convolutional long short-term memory (LSTM) layers between the convolutional backbone and detection head. They compare their results to the nuScenes implementation of PointPillars by using ten $50\ ms$ frames, totalling $500\ ms$, as well as a version using three $50\ ms$ frames, totalling $150\ ms$. Whilst this method still uses full LiDAR sweeps, it does employ the process of using temporally consecutive frames to assist with detection. The use of LSTM layers improves mean average precision (mAP) results in some of their experiments, but it consistently increases the processing time by 3.5 - 4 times.

A transformer model that focuses on the temporal aspect of sequential LiDAR frames is proposed by Yuan \textit{et al.} \cite{tc_trans_lidar}. They use the PointPillars feature extraction layer to generate a pseudo-image before passing this to a feature aggregation and refinement layer which includes the temporal-channel transformer. This approach also leverages temporally consecutive frames, and increases the mAP by $20$ \% over PointPillars on nuScenes. However, the function of the temporal-channel transformer is to learn the interactions between frames, rather than to increase efficiency by only operating on the changing sections of the point cloud. This difference is key, as slow inference speeds due to the inherent complexity of attention-based networks like transformers remain a challenge and area of ongoing research \cite{conv-attention}, \cite{self-attention}. Another example of transformer-based temporal object detection is the Point-Trajectory Transformer, which utilizes long-term, short-term, and future memory modules, as well as a point-trajectory aggregator to predict an object's trajectory and enhance detection across frames~\cite{ptt}. This approach attains similar mAP improvements to the work by Yuan \textit{et al.} \cite{tc_trans_lidar}, however, it still has an inference time of $150\ ms$, even when running on an Nvidia 3090 GPU.

By applying elements of an attention network, specifically a weighting process, Zhong \textit{et al.} aim to enhance PointPillars with SWFNet \cite{swfnet}. A point sampling and weighted fusion of the points, features and pillars occurs prior to scattering in the pillar feature net. When compared on the KITTI 3D detection benchmark this leads to a $4.33\%$ mAP improvement over PointPillars, with a fastest runtime of $8.70\ ms$. However, aggressive sampling to attain speed increases can lead to safety critical issues, especially when detecting extremely sparse targets like drones. Drones flying over structures might be entirely excluded from pillars due to sampling, compromising both safety and performance.

FlowNet3D~\cite{flownet} estimates scene flow from consecutive point cloud frames, and thus can be applied to motion segmentation. SLIM~\cite{slim} is a scene flow and motion segmentation network that builds on this work. The network utilizes a pillar feature encoder, capable of accurately estimating the motion and direction of points in the scene, showing the pillar feature encoders capability of adapting to a motion-based domain. VelocityNet~\cite{velocitynet} is another network which leverages consecutive frames to increase data density, whilst also being influenced by scene flow estimation networks. VelocityNet utilizes a UNet encoder/decoder with a variant of deformable convolutions~\cite{deformable_conv} called time-deformed convolution. Time-deformed convolutions are a form of 3D convolution that apply a time offset as well as a translational one, implementing gather and scatter operations similar to sparse convolutions. Instead of learned positional offsets, the network employs velocity maps, which are birds eye view (BEV) maps giving the direction of moving points in a frame. VelocityNet reports significant 5.7\% mAP improvement over PointPillars in its results when tested on nuScenes. However, this comes at the cost of an almost twenty-fold increase in floating point operations in the network, from 44 billion to 868 billion FLOPS.

\begin{figure*}[hbt]
    \includegraphics[width=7.14in]{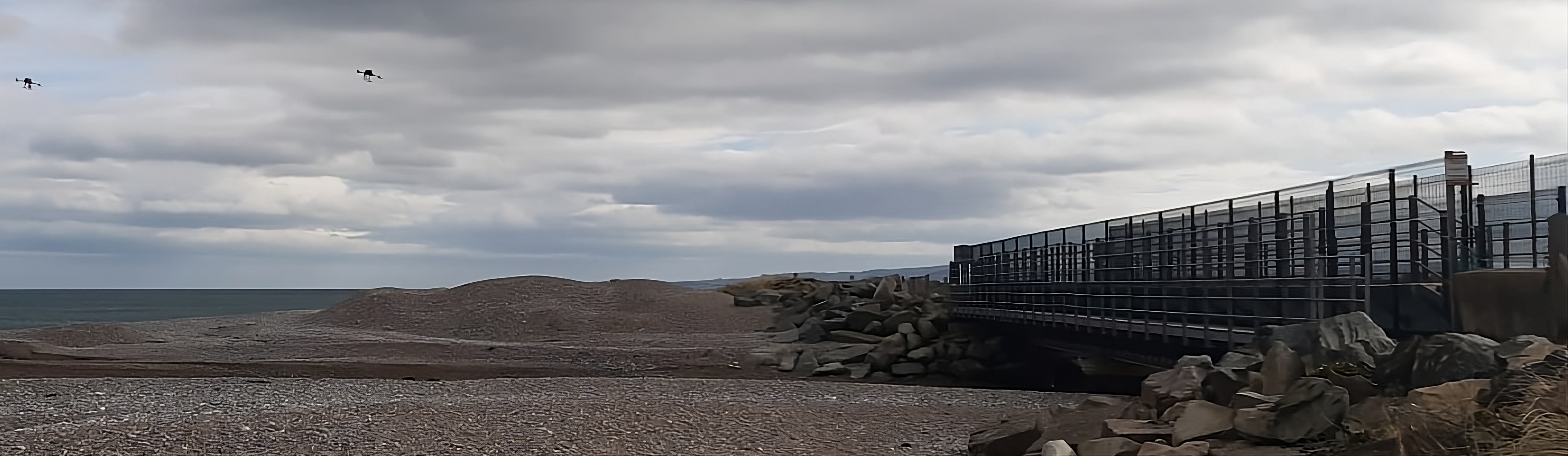}
    \caption{Two DJI M300 RTK drones midflight (top left), with railway bridge (right). The left-most drone is equipped with a Zenmuse L1 LiDAR. The drone to its right is equipped with the Zenmuse P1 photogrammetry camera.}
    \label{fig:2_drones_air}
\end{figure*}

Candidate networks for point-based object detection in safety critical use cases such as drone embedded SAD are required to handle high-throughput in a time sensitive way. Due to both its computational efficiency and high accuracy, PointPillars is a suitable candidate network for modification in this case. The previous work reviewed in this section has focused on increasing the accuracy of PointPillars, but this has come at the cost of speed, or by use of aggressive sampling which cannot be employed in a safety critical use case. Therefore, this work focuses on decreasing latency without affecting accuracy, improving the efficiency of the network to enable faster response to potential hazards on resource restricted embedded computing. To achieve this, we make use of the per-point timestamps available in the data we collect, in contrast to the per-frame timestamps in popular self-driving datasets such as nuScenes, to automatically reuse convolution calculations in unchanged regions of the point cloud. We also leverage the inherent sparsity of point clouds through the use of sparse convolutions, as the idea of sparse convolutions to remove redundant calculations on empty sites has become a much researched topic.

\subsection{Sparse Convolutions}
Graham \cite{graham2014spatially} describes the spatial sparsity of feature maps and how sparsity applies to performance improvements appertaining to different applications. Submanifold sparse convolution \cite{Submanifold} is an efficient sparse convolution operation in which the output will only be computed if the site at the input is active.

Matrix multiplication is an efficient method for implementing spatial convolution on a graphics processing unit (GPU), as demonstrated by Shetlur \textit{et al.} \cite{cudnn-arxiv}. This form of efficient convolution requires transforming the input image into matrices that are suitable for fast multiplication. The transformation can be achieved by lowering the input data with duplication. The GPU implementation of the proposed method is also based on fast GPU matrix multiplication, but without the aforementioned data lowering. 

Work by Gale \textit{et al.} \cite{sparsemm} develops GPU kernels for two sparse matrix operations: sparse matrix-dense matrix multiplication and sampled dense-dense matrix multiplication, which are applied to accelerate the computation of several deep learning models, including MobileNet \cite{mobilenet}. 

Choy \textit{et al.} \cite{minkowski} use sparse tensors and generalized sparse convolutions, which can handle any discrete convolutions in high-dimensional spaces. A sparse convolution with a rulebook is presented in SECOND \cite{second}, where the rulebook implements the mapping between input and output. A highly efficient Sparse Kernel Generator is developed in TorchSparse++ \cite{torchsparse}, which produces high performance sparse point cloud convolution kernels at a fraction of the engineering cost of the current state-of-the-art system.

Vedder and Eaton \cite{sparse_pointpillars} apply the use of sparse tensors and convolutions to PointPillars in the convolutional backbone. To enable more efficient implementation on embedded systems, they replace dense convolutions with sparse and submanifold convolutions and improve the overall runtime two-fold on Nvidia Jetson hardware at the cost of as much as a 9.16\% decrease in mAP. Even this optimization encounters delays in their runtime due to GPU pipelining problems, and the overall runtime for their best scenario on embedded hardware takes $88\ ms$, which means a latency of nearly $200\ ms$ when data acquisition is included.

Shi \textit{et al.} \cite{pillarnet} also apply the use of sparse tensors and convolutions in a point cloud-based object detection network. They follow an architecture of pillar feature extraction to sparse feature extraction backbone to spatial-semantic feature fusion neck to center-based detection head. The network is performs well and is fast, recording 16 FPS on the nuScenes dataset on a Nvidia RTX 3090 GPU, but this is still slower than PointPillars' 62.5 FPS recorded on the nuScenes detection leaderboards.

A commercial drone such as the DJI M300 can travel over $4\ m$ in $200\ ms$, as such we leverage the continuous stream of data that LiDAR outputs to reduce the data acquisition window to $10\ ms$. This results in a maximum allowance of $10\ ms$ for processing, and a total latency of $20\ ms$ to keep within real-time processing parameters. To achieve this, our work further increases the efficiency of sparse convolutions, both as a single layer and within the PointPillars network, by applying our data reuse scheme to sparse scatter-based convolutions, resulting in consistent real-time processing of data in under $10\ ms$. 

\subsection{LiDAR as a Streaming Sensor}
The KITTI dataset~\cite{kitti} and associated benchmark has been a highly influential dataset for autonomous driving and LiDAR-based object detection. A key aspect of the KITTI dataset is that the LiDAR frames are synchronized to RGB camera images, and the $100\ ms$ frames are matched to an RGB image. The popular datasets PandaSet~\cite{pandaset},  Waymo Open Dataset~\cite{waymo} and Argoverse 1~\cite{argoverse} follow this same temporal frame, whereas nuScenes [5] uses $50\ ms$ frames. These datasets abstract the LiDAR point cloud construction process. 

In most cases, this results in the spatial-temporal properties of the LiDAR scanning pattern being lost, and timestamp information for the laser pulse acquisition is not retained. nuScenes, Waymo, KITTI, and Argoverse, to the best of our knowledge, do not provide per-point timestamps. PandaSet is notable for being the only such popular autonomous driving point-based object detection dataset to include per-point timestamps, although is limited to just 8 second long series of sequential point data recording.

One reason for the choice of frame length can be attributed to the number of points needed to accurately detect objects within a scene. Whilst deep learning-based point generation methods such as GgPG~\cite{GgPG} can reduce sparsity and perform well, especially on small objects with sparse point clouds, the processing time is currently prohibitive for use in real-time embedded systems. Deformable Pyramid Voxel R-CNN, by utilizing a deformable voxel ROI pooling method, also performs well when detecting sparse objects ~\cite{deformable_pyramid_rcnn}. However, the authors note their approach is not as efficient as Voxel R-CNN, which in turn is slower than PointPillars, even when utilizing a more powerful GPU~\cite{deng2021voxel}. Due to these limitations in processing time, this work will now consider temporal-focused approaches.

Another reason for the choice of frame length can be the time taken to gather a full $360\degree$ sweep, however as shown in~\cite{strobe},~\cite{han_streaming_2020}, this leads to outdated positional results. The StrObe network~\cite{strobe} and Han \textit{et al.}~\cite{han_streaming_2020} take the approach of taking $10\ ms$ packets of $360\degree$ birds-eye view (BEV) LiDAR sweeps, which return a $36\degree$ wedge. StrObe identifies that the small detection window can make accurate detections difficult, and so employs the use of multi-scale spatial memory maps, which it concatenates with and updates at each convolutional layer.

PolarStream~\cite{polarstream} builds on the work of~\cite{strobe},~\cite{han_streaming_2020} by addressing the computational inefficiency of representing the wedges in a rectangular region, instead using polar coordinates to represent the packet as a wedge with no added noise. PolarStream compares favorably with both~\cite{strobe},~\cite{han_streaming_2020} in terms of mAP.

Abdelfattah \textit{et al.}~\cite{abdelfattah_multi-modal_2023} introduce a multi-modal approach to streaming 3D object detection by replacing the LiDAR slices used by Han \textit{et al.}~\cite{han_streaming_2020} with RGB camera imagery. When tested on NuScenes, they show that this approach has a higher mAP than both PointPillars and PolarStream, with their best framework claiming a 45 FPS runtime.

SpikeClouds~\cite{spikeclouds} is a streaming LiDAR object detection network which utilizes a combination of a spiking neural network (SNN) backbone and a CNN-based neck and detection head. SpikeClouds uses this architecture to process individual scanlines produced by MEMS-based LiDARs, in contrast to previous works which focus on rotational scanners. The SNN processes each scanline individually and the CNN processes the combined SNN output once per scan. As such, backbone processing occurs throughout the acquisition time, and only neck and detection head processing have to wait for the full scan completion. Their work, therefore, still requires a full $100\ ms$ scan before processing, leading to an end-to-end processing time of $120\ ms$ on an Nvidia 2080 GPU. In terms of performance, SpikeClouds achieves an mAP score within $10$ points of PillarNet, and is a notable contribution which broadens LiDAR object detection away from work which focuses solely on rotational scanners.

\begin{table*}[htb]
\caption{Related Work Comparison}
\label{tab:related-work-table}
\begin{tabular}{|p{62pt}|p{135pt}|p{135pt}|p{135pt}|}
\hline
Method & Purpose & Limitations & Solutions Informing SSCATeR's Design\\
\hline
PointPillars~\cite{pointpillars} & Detect objects related to autonomous driving in point clouds. & Network is fast on non-embedded hardware, and utilizes $100\ ms$ frames. Designed for ground-based detection, rather than aerial. & Redesigns the feature extraction and backbone modules to use work in real-time on embedded hardware, using $10\ ms$ frames. Expand the network's detection capabilities along the Z-Axis.\\
\hline 

McCrae and Zakhor~\cite{mccrae_3d_2020} & Utilizes LSTM and sequential point cloud frames to improve mAP over PointPillars. & \multirow{5}{=}[14pt]{\parbox[t]{135pt}{Increased processing time to achieve higher performance.}} & \multirow{5}{=}{\parbox[t]{135pt}{Produces the identical outputs compared to baseline, and so the same performance metrics, whilst implementing a more computationally efficient architecture to enable real-time processing.}} \\
\cline{1-2}

Yuan \textit{et al.}~\cite{tc_trans_lidar} & PointPillars based feature extractor with temporal-channel transformer leveraging sequential frames for improved mAP. &  &  \\
\cline{1-2}

Point-Trajectory Transformer~\cite{ptt} & Transformer-based approach utilizing long-term, short-term, and future memory modules for improved mAP. & & \\
\cline{1-2}

GgPG~\cite{GgPG} & Point generation network which reduces sparsity to improve detection results. & & \\
\cline{1-2}

Deformable Pyramid Voxel R-CNN~\cite{deformable_pyramid_rcnn} & Uses a deformable voxel ROI pooling method to improve detection of sparse objects. & & \\
\hline

SWFNet~\cite{swfnet} & Applies a sampling algorithm in the PointPillars-based featuring extraction module to attain higher mAP and faster inference. & Aggressive sampling of points risks sampling out safety critical objects such as extremely sparse drones. & Does not apply sampling of safety critical points to the input. \\
\hline

VelocityNet~\cite{velocitynet} & Applies scene flow and time-deformed convolutions to track points and assist in increasing the mAP score. & Introduces a near twenty-fold increase in floating point operations over the PointPillars network. & Focuses on maintaining performance metrics whilst reducing operation count for a more lightweight design. \\
\hline

Sparse PointPillars~\cite{sparse_pointpillars} & Use sparse convolutions to reduce the inference time of PointPillars for real-time embedded processing. & Utilizes $100\ ms$ frames, meaning that real-time processing in their use-case still creates a $200\ ms$ delay. & \multirow{2}{=}{\parbox[t]{135pt}{Also follows a sparse approach, but utilizes temporally sequential frames and data reuse to further reduce computation.}} \\
\cline{1-3}

PillarNet~\cite{pillarnet} & Use sparse convolutions and a powerful feature encoder to address the gap between pillar- and voxel-based point cloud object detection & Utilizes $100\ ms$ frames, meaning that real-time processing in their use-case still creates a $162.5\ ms$ delay. & \\
\hline

Han \textit{et al.}~\cite{han_streaming_2020} & Slice $360\degree$ LiDAR sweeps into subsets and use LSTM and stateful NMS to track objects and reduce latency. & Latency reductions come at the cost of mAP. Relies on $360\degree$ LiDAR sweeps. & Produces identical outputs to the baseline, so latency reductions do not impact performance. Works with non-repetitive LiDAR scanning patterns. \\
\hline

StrObe~\cite{strobe} & Slice $360\degree$ LiDAR sweeps into subset packets and processing with multi-scale spatial memory and HD maps to reduce latency while improving mAP. & \multirow{2}{=}{\parbox[t]{135pt}{Latency reductions still perform above real-time, even when using non-embedded, desktop hardware. Relies on $360\degree$ LiDAR sweeps.}} & \multirow{2}{=}{\parbox[t]{135pt}{Can achieve real-time processing even on embedded hardware. Works with non-repetitive LiDAR scanning patterns.}} \\
\cline{1-2}

PolarStream~\cite{polarstream} & Slice $360\degree$ LiDAR sweps into subset packets and processes them using polar coordinates to improve efficiency and mAP results. & & \\
\hline

Abdelfattah \textit{et al.} ~\cite{abdelfattah_multi-modal_2023} & Uses a multi-modal approach combining RGB and LiDAR sliced subset packets to improve efficiency and mAP results. & Latency reductions still perform above real-time, even when using non-embedded, desktop hardware. Relies on $360\degree$ LiDAR sweeps. Requires two sensors which increases weight and reduces drone flight efficiency. & Can achieve real-time processing even on embedded hardware. Works with non-repetitive LiDAR scanning patterns. Does not require multiple sensors.\\
\hline

SpikeClouds~\cite{spikeclouds} & Processes LiDAR scanlines through an SNN backbone during acquisition, meaning only neck and detection head processing must wait for the full LiDAR scan before processing. Is able to work with non-rotational LiDAR scanners. & Requires a full LiDAR scan before detections can be processed. & Does not require a full LiDAR scan to complete detections. Also works with non-rotational LiDAR scanners. \\
\hline

\end{tabular}
\end{table*}

In this paper we use a single LiDAR sensor for scene perception due to the size, weight, and power constraints of the drone paradigm. We leverage the PointPillars pillar grid to capture any points entering or leaving the grid during every $10\ ms$ interval. Subsequently, by sending only those pillars that have changed to the convolution layer and by retaining the result of previous convolutions within the network, we achieve efficiency by recycling rather than recalculating the convolutions on unchanged pillars during each forward pass. We summarize the comparisons of our approach to existing methods in Table \ref{tab:related-work-table}.

\begin{figure}[tb]
    \includegraphics[width=2in, center]{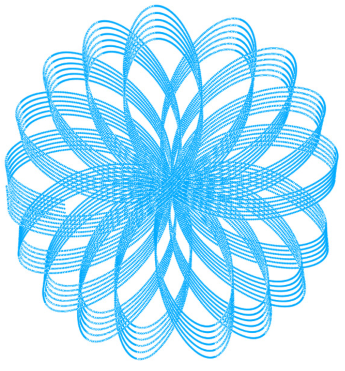}
    \caption{Non-Repetitive Scanning Pattern of the DJI Zenmuse L1 over $100$ milliseconds.}
    \label{fig:avia_pattern}
\end{figure}
\begin{figure}[bt]
    \includegraphics[width=3.49in]{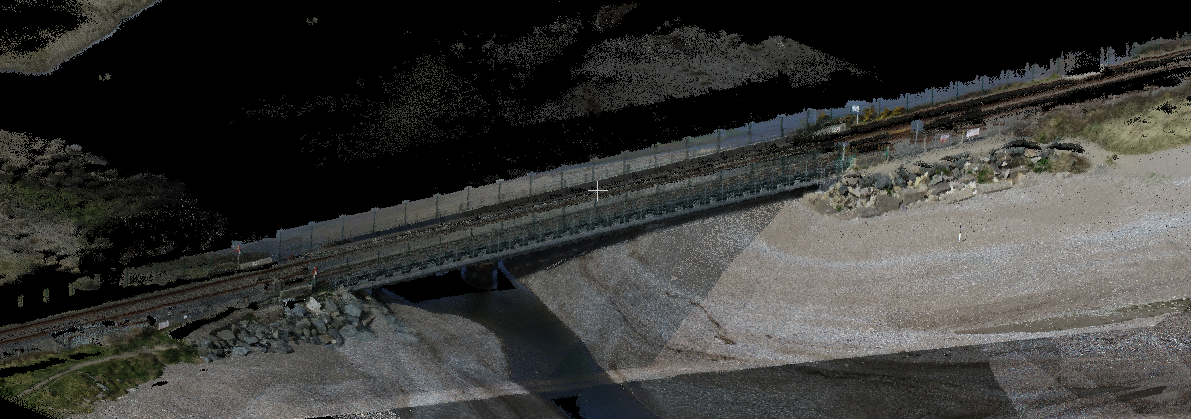}
    \caption{A true-color point cloud over the over-water railway bridge collected in Newcastle, County Wicklow, Ireland.}
    \label{fig:nc_pointcloud}
\end{figure}

\section{Methodology}
\label{sec:methodology}

In this section, we first outline our dataset collection and labeling process and equipment. We describe PointPillars and the modifications we have made, including the data reuse scheme. Finally, we outline the testing parameters in comparison to sparse convolutions. Although we use PointPillars, the method can be adapted to other point-based object detection networks that employ a feature extraction layer and a convolutional backbone.

\subsection{Dataset Collection}
\label{sec:datasets}

\begin{figure}[thb]
    \includegraphics[width=3.49in]{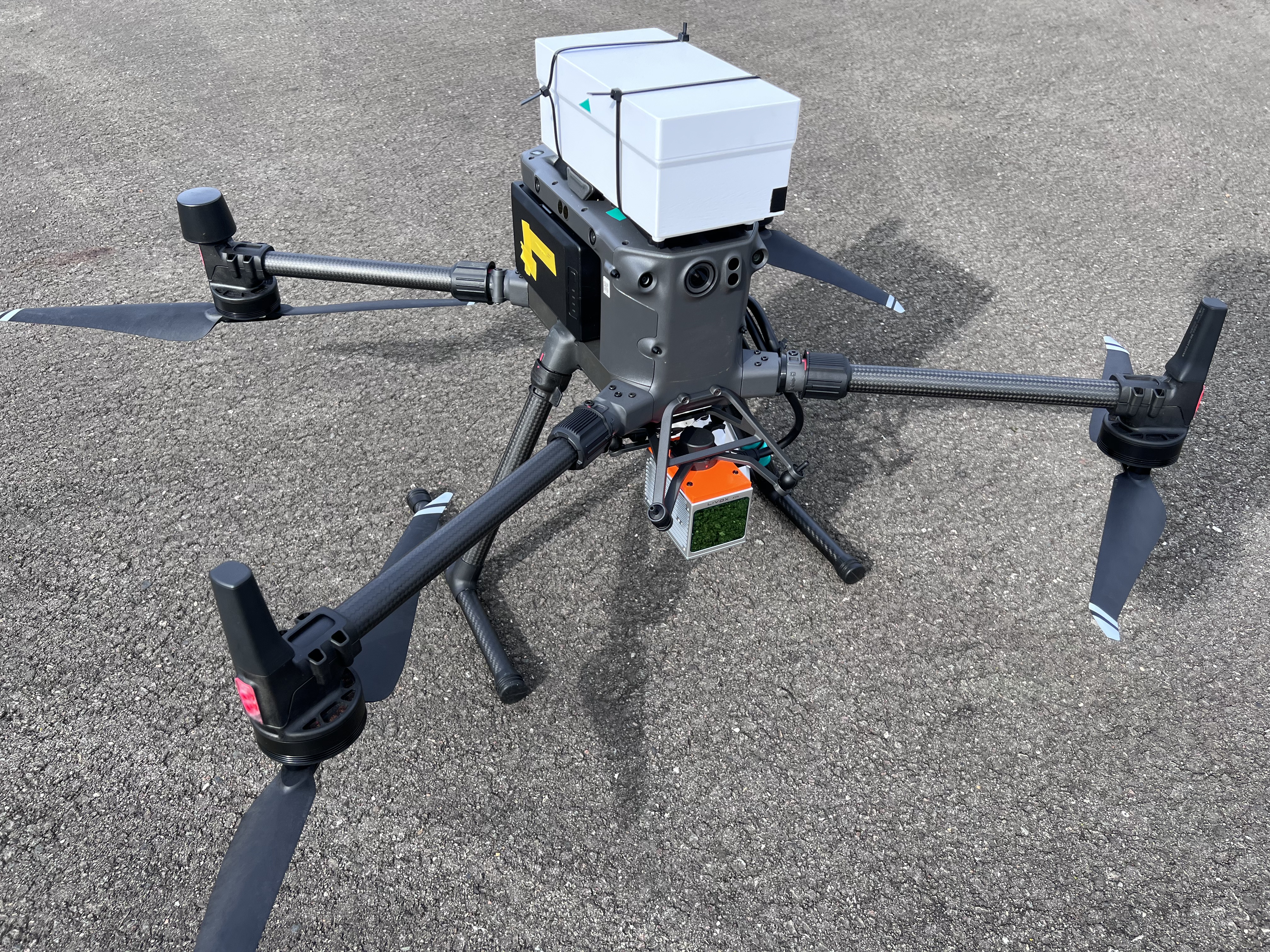}
    \caption{Livox Avia mounted below a DJI M300 RTK drone, with embedded computer system mounted on top.}
    \label{fig:livox_setup}
\end{figure}

To enable model training and testing, we collected multi-drone survey datasets at two contrasting locations in Ireland and Germany. The first location is a railway bridge in Newcastle, County Wicklow, Ireland \cite{denoise}. Two DJI M300 drones are utilized as shown in Fig. \ref{fig:2_drones_air}. One drone carried the DJI Zenmuse L1 LiDAR system as its payload for data collection. The five channel L1 LiDAR utilizes a non-repetitive scanning pattern enabled by the use of Risley prisms as shown in Fig. \ref{fig:avia_pattern} for a $70.4\degree \times 77.2\degree$ field-of-view (FoV) with a pulse rate of $240,000$ points per second. The reconstruction of the collected survey point cloud is displayed in Fig. \ref{fig:nc_pointcloud}. The DJI L1 system, however, cannot be used for real-time in-field testing, as the data is not accessible in real-time. As such, we have created an onboard compute system which utilizes the Livox Avia, the LiDAR unit found inside the L1, which is pictured in Fig. \ref{fig:livox_setup}.

\begin{figure}[tb]
    \includegraphics[width=3.49in]{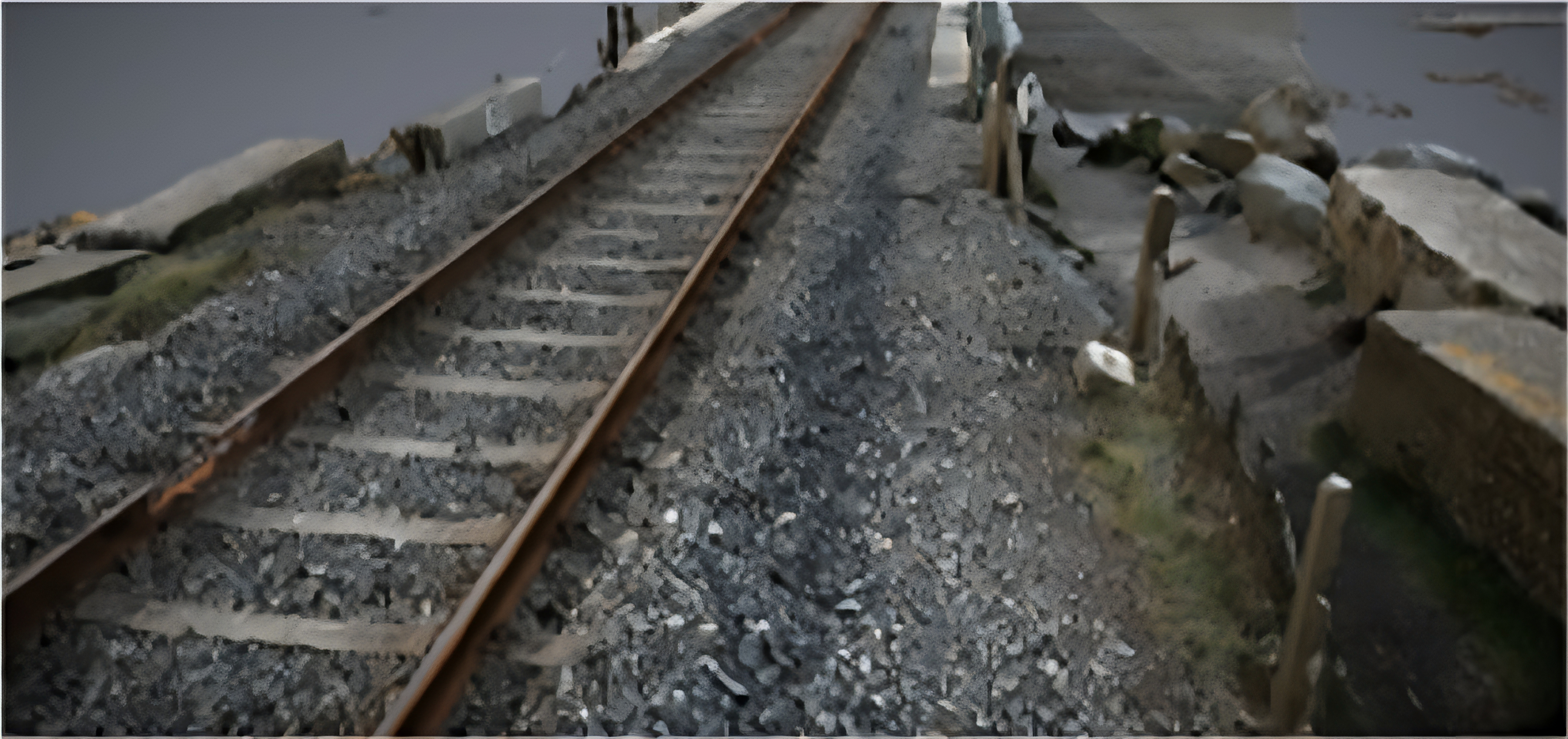}
    \caption{Close-range photogrammetry of the railway bridge.}
    \label{fig:track_photo}
\end{figure}

The second drone was equipped with a DJI Zenmuse P1. It performed a photogrammetry survey, producing the 3D model shown in Fig. \ref{fig:track_photo}. Typically, LiDAR missions are conducted at greater target range than photogrammetry missions (eg. $50\ m$ vs $10\ m$) as photogrammetry requires higher ground sample density for accurate reconstruction. The spatial separation of the drone flight paths allow for simultaneous operation of both missions. However, the LiDAR equipped drone frequently observes returns from the photogrammetry drone over the course of a survey mission. The estimated position of the drones was measured using real-time kinematic (RTK) GNSS and used to annotate drone observations in the LiDAR point cloud dataset to cm-level accuracy. DJI TimeSync enables time synchronization in each drone, and utilizes the RTK source, which is the same for all operating drones.

The second scenario was the Harburg Lock, in the Port of Hamburg, Germany \cite{manduhu2023airborne}. Three DJI M300 drones are used. Two drones simultaneously inspected for cracks in the lock walls using their onboard cameras. The two camera drones were required to capture imagery of the Lock wall with c. 1 mm Ground Sample Density and thus flew within the Lock chamber within 1-2 m of the Lock wall. While their flight plans were strategically separated using trajectory-based planning, their flight plans risked denial of GNSS and communications due to multi-path and signal occlusion by the walls of the Lock chamber. A third drone equipped with a Livox Avia with SAD embedded compute or DJI L1 LiDAR system, dependent on data collection or testing, was deployed to monitor the two camera drones. The LiDAR drone was positioned to station keep at one end of the Lock and at a flight elevation midway between the two camera drones. The SAD algorithm onboard the LiDAR drone was configured to raise an alert if the 15 m separation threshold was breached, which was communicated to the remote pilot via the command and control (C2) radio link.

The datasets are summarized in Table \ref{datasets_table}, where the $10,000$ frames refer to $100\ ms$ frames that increment in $10\ ms$ time intervals.

\begin{table}
\caption{Summary of Collected Datasets} 
\label{datasets_table}
\setlength{\tabcolsep}{3pt}
\begin{tabular}{|p{32pt}|p{25pt}|p{50pt}|p{43pt}|p{70pt}|}
\hline
Location&Total Frames&Average Points Per Frame&Points Range Per Frame&Environment and Scenario\\
\hline
Newcastle&$10,000$&$5798.48$&$[781, 7806]$&Natural environment bridge inspection.\\
\hline
Hamburg&$10,000$&$5546.53$&$[690, 6270]$&Built environment port lock wall inspection.\\
\hline
\end{tabular}
\end{table}

\begin{figure*}[thb]
    \includegraphics[width=7.14in]{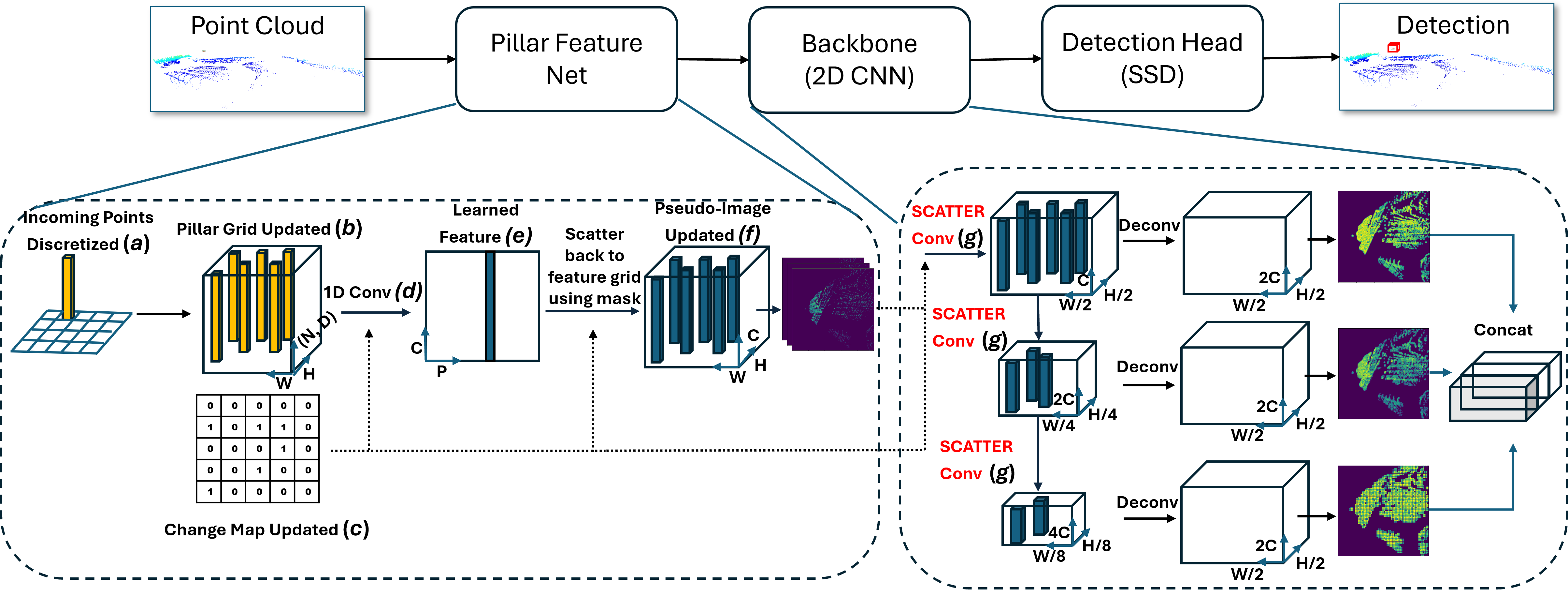}
    \caption{Network overview of the PointPillars architecture. We modify the architecture to include change maps which inform the network where to perform convolutions. In the convolutional backbone we utilize SCATTER in place of standard sparse convolution to increase processing efficiency.}
    \label{fig:pp_arch}
\end{figure*}

\subsection{PointPillars Architecture and Modifications} \label{pp_arch_method}

The PointPillars network comprises three primary components: the Pillar Feature Net (PFN), the convolutional backbone layer, and the detection head. These components are shown in Fig. \ref{fig:pp_arch} along with our modifications and visualizations of different component outputs.

\begin{figure}[bth]
    \includegraphics[width=3.49in]{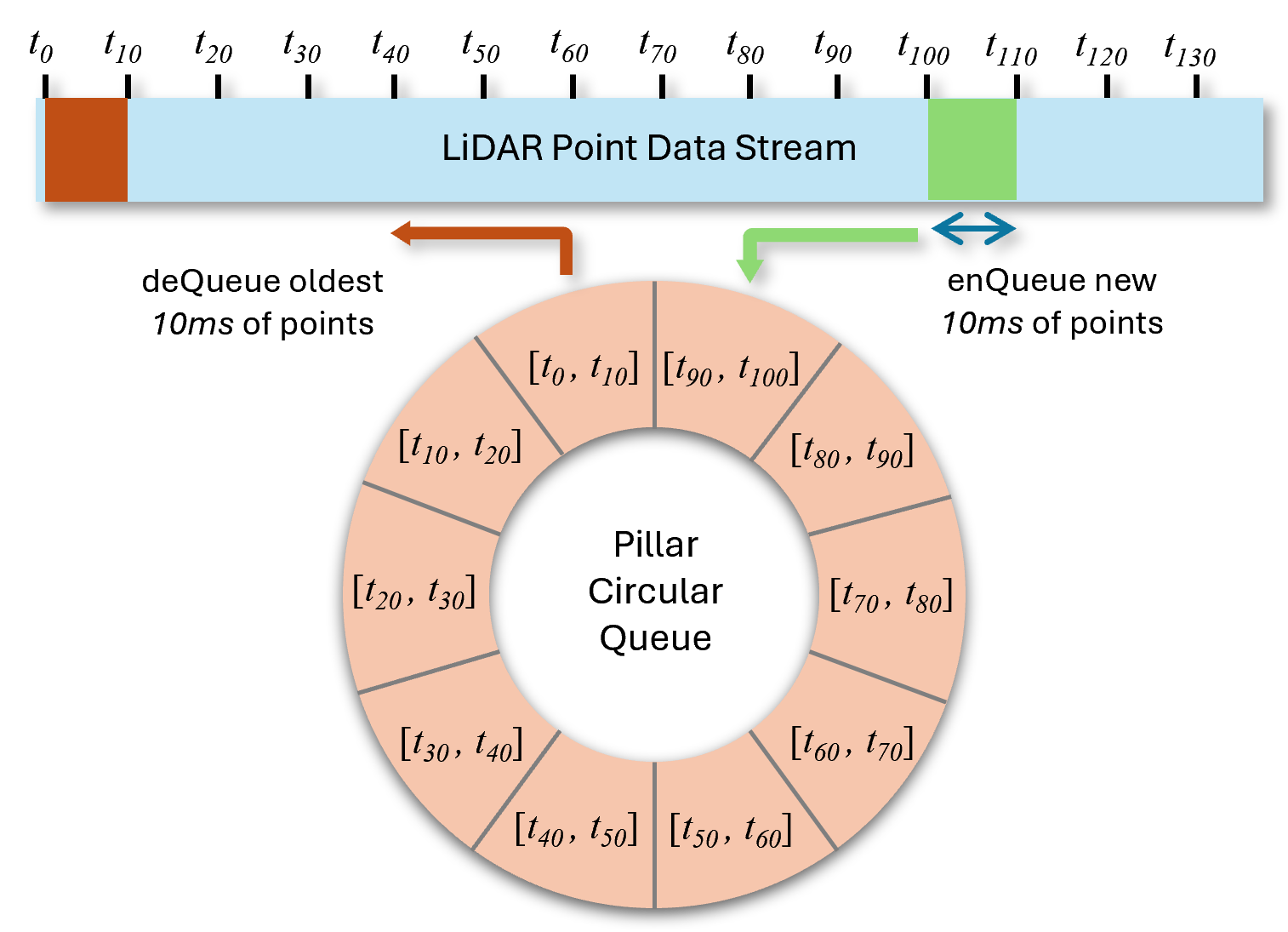}
    \caption{A graphic of the pillar input and output process with $130\ ms$ LiDAR point data stream (top).}
    \label{fig:pillar_diagram}
\end{figure}

The PFN receives raw 3D point cloud data as input and organizes it into pillars (Fig. \ref{fig:pp_arch} (a)), where each pillar is a vertical stack of $N$ number of points, where $N = 100$. If a pillar contains less than $N$ points it is zero-padded, to provide a homogeneous input to the neural network. The number of pillars, $P$, are set at $12,000$, which are also zero-padded when there are less than $P$ pillars. 

The original PointPillars architecture uses random sampling in case of overflow. We use the original parameters for $N$ and $P$, and incorporate an assertion to catch any instance of random sampling to ensure that hazard-related points are not being filtered out. The random sampling of points can lead to false negatives, as mentioned previously, and thus, an increased risk of collision.

The pillar grid is updated with the new points, as shown in Fig. \ref{fig:pp_arch} (b). In our implementation, change maps (Fig. \ref{fig:pp_arch} (c)) are used to track which pillars have been modified in the previous $10\ ms$ due to insertion or removal of points, as shown in Fig. \ref{fig:pillar_diagram}. The processing of the pillars in the PFN is treated as a circular queue. The most recently acquired points are inserted by enQueue operation while any points older than $100\ ms$ are removed by deQueue and converted. Each 2D change map tracks the updated pillars to inform the subsequent CNN backbone layers where to recalculate convolutions for the changed sites and where the previous convolution results can be reused for unchanged sites.

After creating the pillars, 1D convolutions (Fig. \ref{fig:pp_arch} (d)) are performed on changed pillars, producing a $(C, P, N)$ tensor, where $C = 64$ giving $64$ channels. A max operation is then performed across the channels leaving a $(C, P)$ tensor. The resultant learned features (Fig. \ref{fig:pp_arch} (e)) are then scattered back to the feature grid to create the pseudo-image input to the convolutional backbone (Fig. \ref{fig:pp_arch} (f)), along with the associated change map.

The original PointPillars architecture used pillar $x$, $y$ horizontal dimensions of $0.16\ m$ when applied to the KITTI dataset. In our implementation we retain the $0.16\ m$ pillar horizontal dimensions, as from a birds-eye view perspective, the cross sectional dimensions of the smallest labeled object in KITTI, pedestrians, and our multi-rotor drones are similar. In the KITTI dataset the majority of pedestrian lengths and widths are between $0.5\ m$ and $1.0\ m$ \cite{kitti_dims}. The DJI M300's length and width from outstretched arm to outstretched arm is $0.65\ m$ and $0.72\ m$ respectively. The real-world detection area is then discretized into a grid of size $W \times H$. For example, given an $80.64\ m \times 80.64\ m$ detection area, this discretizes into a grid of $504 \times 504$ pillars. 

In terms of complexity, in the traditional pillar assignment of points, the complexity is linear dependent on the number of points. This gives us O(n) where n represents the discretization of each point into a pillar. In our implementation, the pillar discretization stage is slightly more complex, as the process involves the addition and removal of points to pillars. However, our number of points needing to be processed is much smaller. We represent this additional linear scaling cost related to the number of pillars as m, while representing our discretization cost as k. This therefore gives our algorithm the complexity O(k + m).

Points are decorated with $D$ feature dimensions, where $D = 9$ dimensions: the spatial coordinates $x$, $y$ and $z$; intensity; the pillar centre offset values $x_p$ and $y_p$; and the distance to the arithmetic mean of the pillar $x_c$, $y_c$ and $z_c$. The pseudocode shown in Fig. \ref{pillar_pseudocode} describes the point removal and insertion process in further detail. The circular queue is shown in Fig. \ref{fig:pillar_diagram}, while the change map is initialized as a boolean array of $504 \times 504$ with each value corresponding to a pillar, and the boolean value denoting whether a pillar is changed or unchanged. A pillar is classed as changed, and therefore reflected with a 1 in the change map, if any points in the pillar are added or removed. New points take six operations to add to a pillar and removing points takes thirteen operations. Further, changed pillars incur an operation cost of $n - 1 + 1$ to calculate the arithmetic mean for all points in the pillar where $n$ is the number of points. Finally, each point in each changed pillar takes three operations to find the $x, y, z$ distance from this arithmetic mean. The algorithm in Fig. \ref{pillar_pseudocode} describes how the change map is modified in the PFN, which occurs between stages (a) to (f) in \ref{fig:pp_arch}, and the Backbone, which occurs from stage (g) onwards in the backbone inset of \ref{fig:pp_arch}.

\makeatletter
\newcommand{\removelatexerror}{\let\@latex@error\@gobble}
\makeatother

\begin{figure}[!t]
 \removelatexerror
\begin{algorithm}[H]
\renewcommand{\algorithmicrequire}{\textbf{Input:}}
\renewcommand{\algorithmicensure}{\textbf{Output:}}
\algnewcommand\algorithmicinit{\textbf{Initialize:}}
\algnewcommand\Initialize{\item[\algorithmicinit]}
\newcommand{\pluseq}{\mathrel{+}=}
\caption{Change Map Initialization and Modification}\label{pillaralg:cap}
\begin{algorithmic}[1]
\raggedright
\Initialize{Initialize circular queue for each pillar \newline
            Initialize pillar grid by building up $100\ ms$\newline
            of data in $10\ ms$ increments \newline
            Initialize change map as $504$ x $504$ array with values set to $0$ \newline
            Initialize $t$ as time}
\Require {$10ms$ increments of streamed LiDAR data}
\Ensure {Pseudo-image of extracted features} \newline
\textbf{Pillar Feature Net}
\For {each $10ms$ streamed LiDAR data}
    \For{each point with timestamp \textless\ t}
        \State {find $\lfloor x, y \rfloor$ of point}
        \State {change map array = $1$ at that location}
        \State {convert each point dimension to zero-padding}
        \EndFor
    \For{each point in new $10ms$ data stream}
        \State {find $\lfloor x, y \rfloor$ and enQueue to pillar}
        \State {change map array = $1$ at that location}
        \State {find point $x_p, y_p$ offset to pillar centre}
        \EndFor
    \For{each pillar at change map array = $1$}
        \For {each point in pillar}
            \State {find $x_c, y_c, z_c$ distance to pillar \hspace*
            {\algorithmicindent}\hspace*{\algorithmicindent}\hspace*{\algorithmicindent}arithmetic mean}
            \EndFor
        \EndFor
    \State {Increment $t$ by 0.01}
    \State \parbox[t]{\dimexpr\linewidth-\algorithmicindent}{Pass to 1D Conv layer for feature extraction and produce pseudoimage}
\EndFor
\textbf{Backbone}
\For {each pseudoimage}
    \If{change map array = $1$}
        \State{Apply sparse or sub-manifold convolution and\newline \hspace*
            {\algorithmicindent}\hspace*
            {\algorithmicindent}scatter results to associated neighboring output \hspace*
            {\algorithmicindent}\hspace*
            {\algorithmicindent}sites}
        \If{output site is changed by convolution or\newline\hspace*
            {\algorithmicindent}\hspace*
            {\algorithmicindent}ReLu} 
            \State{change map array = $1$ at that site}
        \EndIf
        \If{convolution stride \textgreater\ $1$ and any site under\newline\hspace*
            {\algorithmicindent}\hspace*
            {\algorithmicindent}kernel has corresponding change site = $1$}
            \State{corresponding downsampled change map\newline\hspace*
            {\algorithmicindent}\hspace*
            {\algorithmicindent}\hspace*
            {\algorithmicindent}site = $1$}
        \EndIf
        \If{deconvolution stride \textgreater\ $1$ and any site under\newline \hspace*
            {\algorithmicindent}\hspace*
            {\algorithmicindent}kernel has corresponding change site = $1$}
            \State{all corresponding upsampled change map\newline\hspace*
            {\algorithmicindent}\hspace*
            {\algorithmicindent}\hspace*
            {\algorithmicindent}sites = $1$}
        \EndIf
    \EndIf        
\EndFor
\end{algorithmic}

\end{algorithm}

\caption{Pseudocode describing the pillar creation and change map initialization and modification process throughout the pillar feature net and backbone.}
\label{pillar_pseudocode}

\end{figure}

The backbone layer, implemented as a convolutional neural network (CNN), processes the pseudo-images and change map. This layer extracts high-level semantic features aimed at capturing the spatial and semantic characteristics of the objects within the point cloud data. The CNN backbone layer is formed of three blocks over two subnetworks, a downsampler followed by an upsampler. Each block is characterized as Block($S, L, F$), where $S$ is the stride length, $L$ is the number of layers and $F$ is the number of output channels. Each block of varying parameters is first downsampled, and then upsampled to be concatenated and passed to the detection head for 3D object detection. In our backbone, Fig. \ref{fig:pp_arch} (g), a SSCATeR convolution is applied at each downsampling stage to mitigate computation time. This convolution is specifically applied to changed sites only, optimizing efficiency without compromising accuracy.

\subsection{Sparse Scatter-Based Convolutions with Temporal Data Recycling: SSCATeR}

\begin{figure}[bth]
    \includegraphics[width=3.49in]{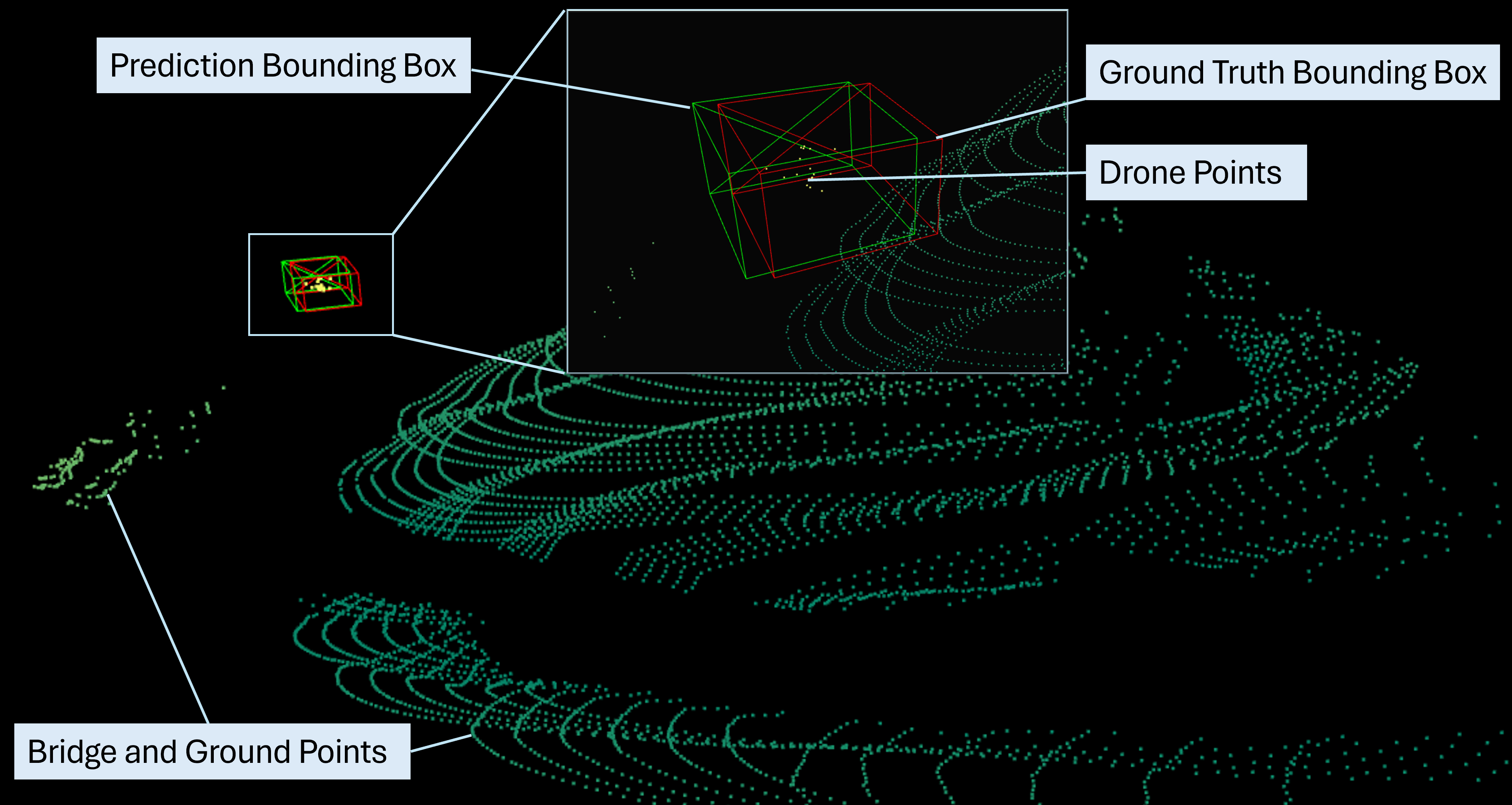}
    \caption{An example prediction of of a drone detection using the Aerial PointPillars network~\cite{manduhu2023airborne} in Newcastle, Ireland. The ground truth bounding box is in red, and the prediction bounding box is in green, the drone points are yellow, whilst the ground points are teal.}
    \label{fig:nc_pred}
\end{figure}

Our scatter-based sparse convolution has been described in our previous work \cite{manduhu2023airborne}, in which the precision and inference time of several 3D detection models are compared against our own, showing that the network's precision is comparable and that the inference time is up to ten times faster. An example visualization of the testing can be seen in Fig. \ref{fig:nc_pred}. We extend its use here to incorporate a data reuse scheme and further improve the inference speed for real-time use on embedded hardware. 

In a standard convolution, each neighboring pixel is multiplied by the corresponding filter weight and added to the output position. In our Convolution with Scatter Operation approach, each pixel is multiplied by different filter weights and added to the different output positions. As we also take into account the temporal aspect of short-stride sequential frames of LiDAR data, we refer to the operation as \textbf{S}parse \textbf{S}catter-based \textbf{C}onvolution \textbf{A}lgorithm with \textbf{Te}mporal Data \textbf{R}ecycling \textbf{SSCATeR}. 

The addition of a deconvolution process is an important aspect of our data reuse scheme as described in section \ref{sec:data-reuse}. We define deconvolution with scatter operation as follows:
\begin{equation}
\begin{split}
O_{i - m +  \floor {k/2}, j - n + \floor {k /2}} \; -= I_{i,j} \times W_{m, n} \\
m \in \{0, ..., (k-1)\}, n \in \{0, ..., (k-1)\} \\
i \in \{0, ..., (p-1)\}, j \in \{0, ..., (q-1)\}
\end{split}
\end{equation}

Where k represents the dimensions of the kernel, m and n represent the indices of that same kernel, and p and q represent the size of the input image.

\begin{figure*}[htb]
    \centering
    \includegraphics[width=7.14in]{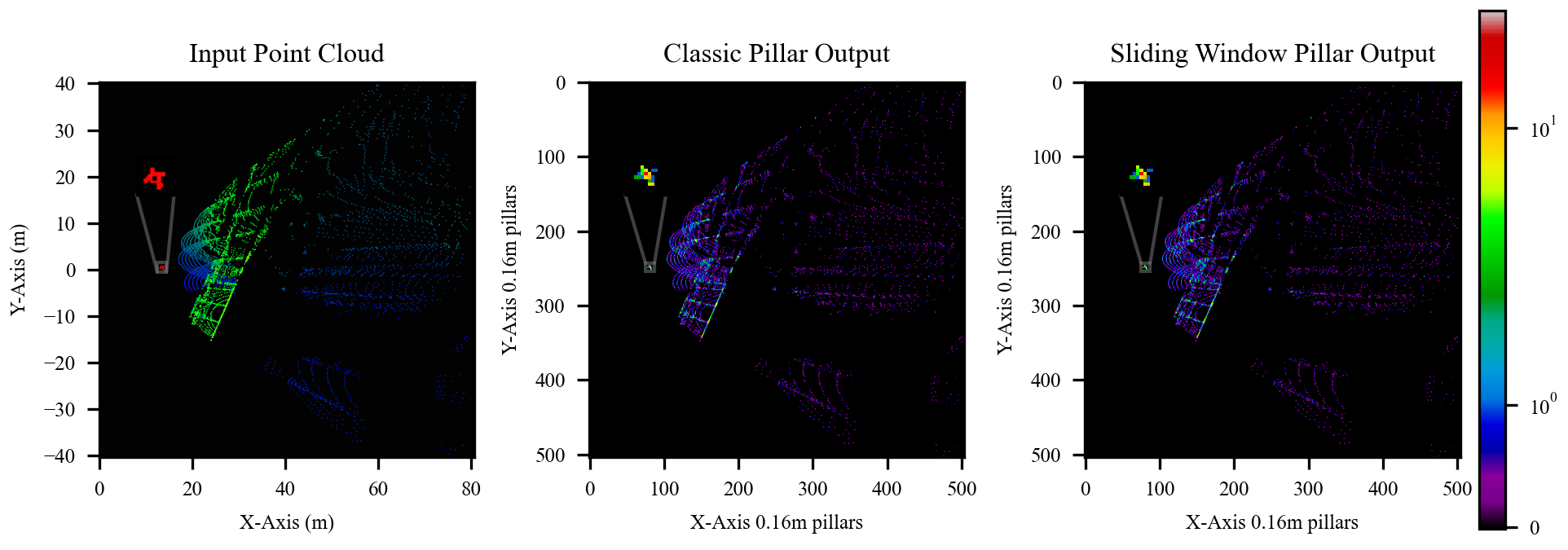}
    \caption{The initial test point cloud (left), collected in Newcastle via drone-mounted LiDAR, cropped to an $80.64m^2$ grid. Discretization of the point cloud yields the same result in both classic (centre) and our method (right).}
    \label{fig:nc_points_and_pillars}
\end{figure*}

\begin{figure*}[htb]
    \includegraphics[width=7.14in]{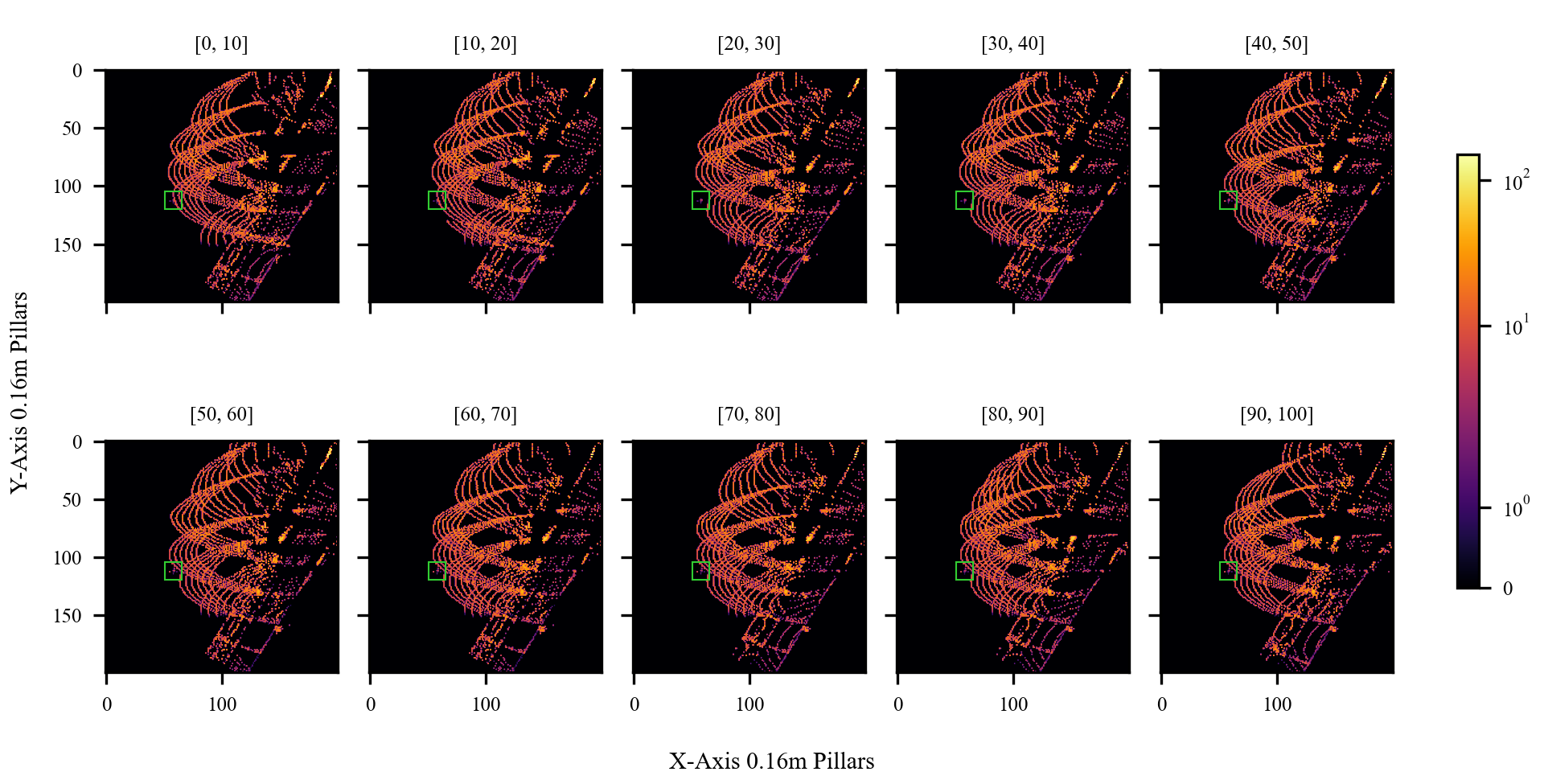}
    \caption{A time series of pseudo-images in $10\ ms$ increments over a $100\ ms$ interval. In this example, the pillars brightness represents the maximum intensity of the points in the pillar. The location of the drone is marked using a green box in each $10\ ms$ increment.}
    \label{fig:nc_pillars}
\end{figure*}

\begin{figure*}[htb]
    \includegraphics[width=7.14in]{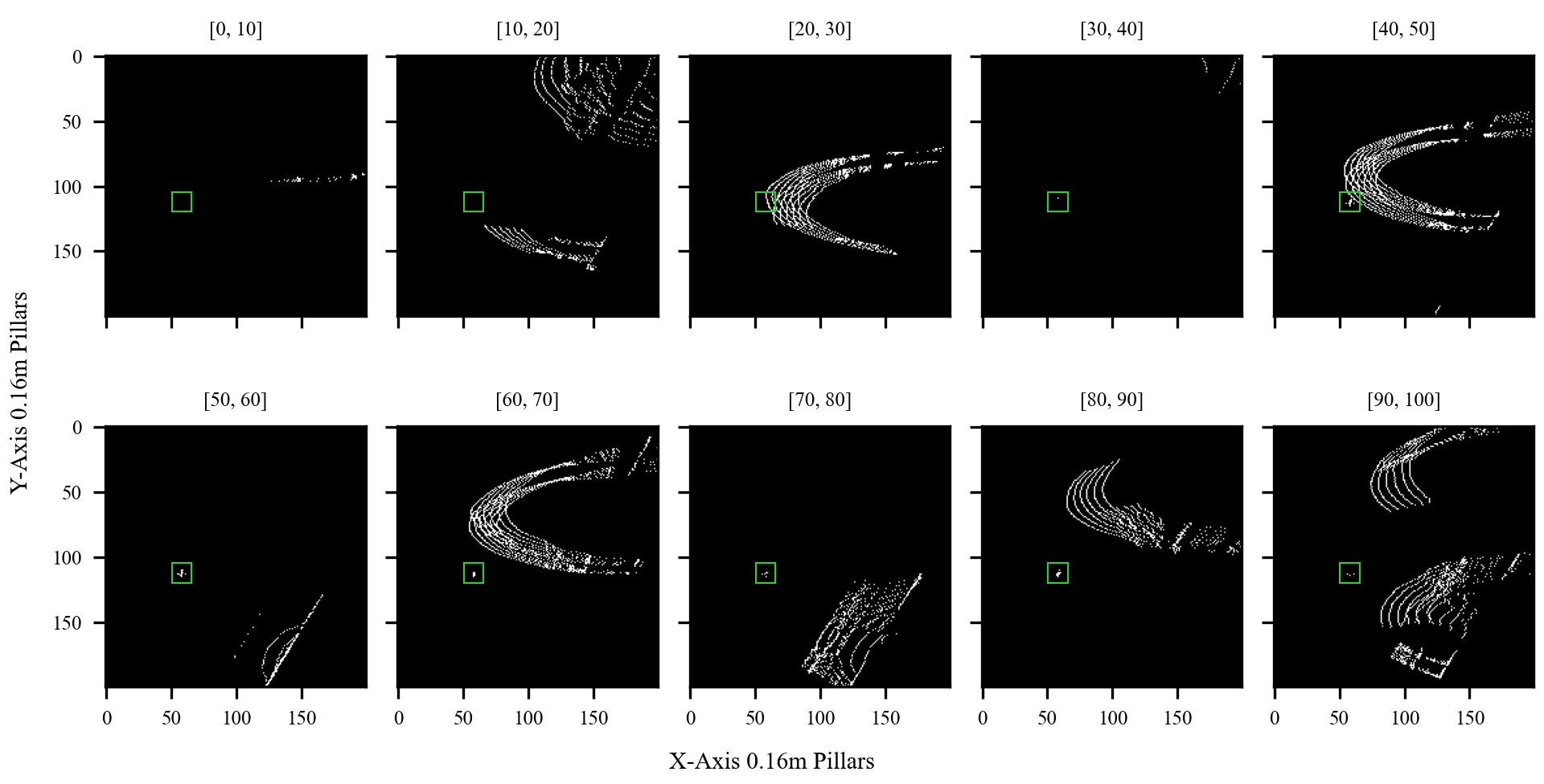}
    \caption{The $10\ ms$ increment binary change maps of the pseudo-images seen in Fig. \ref{fig:nc_pillars}. The location of the drone is marked using a green box in each $10\ ms$ increment.}
    \label{fig:nc_change}
\end{figure*}

\subsection{Data Reuse in Convolution}\label{sec:data-reuse}

We use a subset-based approach to process the input data, which consists of a sequence of LiDAR point cloud frames. Instead of processing the entire frame at each time step, we only update the regions that have changed from the previous frame. This allows us to reduce the computational and memory requirements of our method.

Our method leverages the temporal coherence between the frames by reusing the previous convolution results on unchanged sites. We use a scatter operation to distribute the convolution values to the corresponding output locations.

However, before we can perform the scatter operation, we need to remove the effects of the previous input values on the output. To do this, we use a deconvolution operation to reverse the scatter operation and subtract the values that were previously added to the output locations. This ensures that the output is consistent with the current input and does not contain any artifacts from the previous input. Consequently, we are required to run convolution twice: one run for deconvolution and another run for convolution. In practice we can combine both deconvolution and convolution into one convolution with scatter operation as follows:
\begin{equation}
\begin{split}
O_{i - m +  \floor {k/2}, j - n + \floor {k /2}} \; += (I^{curr}_{i,j} - I^{prev}_{i,j}) \times W_{m, n} \\
m \in \{0, ..., (k-1)\}, n \in \{0, ..., (k-1)\} \\
i \in \{0, ..., (p-1)\}, j \in \{0, ..., (q-1)\}
\end{split}
\end{equation}
where $I^{curr}$ and $I^{prev}$ represent the current input feature and the previous input feature, respectively.

During inference, the value of each change site is multiplied by the filter kernel and the results scattered to the associated neighboring output locations, and the change map is the updated with any newly changed sites. As batch normalization parameters are fixed during inference, it is applied independently at each change site and therefore it does not make any alterations to the change map array. However, the ReLU operation generates zero value when the value is below zero. As a consequence, this generation results in the ReLU operation modifying the change map array, as such we if the ReLU operation generates a zero where there was no zero previously, the change map is set to 1 at that site, to inform that the value there cannot be reused.

Lastly, the detection head operates on the features extracted by the backbone layer, generating a set of predicted 3D bounding boxes for each object detected within the point cloud data. Experiments \cite{wang_yang_optimization_2021} show that the backbone layer dominates the computation, as it comprises multiple CNN blocks. In some state-of-the-art implementations \cite{openpcdet2020}, a sparse convolution approach is utilized to expedite the backbone layer, exploiting the sparsity of the point cloud.

We modify PointPillars' SSD detection head, specifically focusing on the use of anchor boxes. The original PointPillars used anchor boxes for detection, and by default the z-center of these boxes is set to a small $[-1, 3]$ range, reflecting the lateral requirements of autonomous driving. To better fit the network to an aerial domain, we have adapted this section by horizontally stratifying the LiDAR FoV into layers of $1\ m$ height. We allocate an anchor box and corresponding class to each layer.

\subsection{Comparison with a Sparse Backbone and OpenPCDet models}\label{sec:sparse-backbone}

We compared SSCATeR to two network architectures, Aerial-PointPillars and PillarNet, and a total of four implementations of those architectures. Latency is compared between SSCATeR and the four comparison implementations. Performance is compared between Aerial-PointPillars and the two PillarNet implementations, as the weights can be re-used in all implementations of Aerial-PointPillars regardless of whether a rulebook-based sparse, scatter-based sparse or SSCATeR backbone is used.

The original PointPillars network applies dense 2D convolutions in the backbone. Considering the sparsity of the 2D feature maps, sparse and sparse submanifold convolutions make a more challenging, but less biased, benchmark to test SSCATeR against. With this in mind, we have applied both sparse and submanifold convolutions to the PointPillars backbone.

Each block contains the same number of convolutions as per the original PointPillars \cite{pointpillars}, Sparse PointPillars \cite{sparse_pointpillars}, and our previous work's \cite{manduhu2023airborne} parameters in both the Aerial-PointPillars backbones and our SSCATeR implementation. We follow a similar backbone structure as Sparse PointPillars, which uses a standard sparse 2D convolution at the input of each block, followed by submanifold sparse convolutions. The output of each block is then upsampled using transpose sparse 2D convolution and concatenated. For comparison, we utilize both a rulebook-based sparse and submanifold backbone which uses the spconv library~\cite{spconv2022} implementation, as well as our previous work's sparse convolution with scatter operation implementation, which we will refer to as scatter-based sparse convolution. Both Aerial-PointPillars implementations  and SSCATeR use $0.16\ m$ pillars with a grid area of $[0\ m,\ 81.92\ m]$ and $[-40.96\ m,\ 40.96\ m]$ on the $x$ and $y$ axes respectively.

In our SSCATeR implementation, we apply SSCATeR throughout each convolution of each block, both the initial sparse convolution and submanifold convolutions, updating the change maps throughout. We also apply SSCATeR's change maps to the transpose sparse 2D convolution process. 

We implement two PillarNet methods, the first using $0.1\ m$ pillars, as used in the original paper~\cite{pillarnet}, with a grid area of $[0\ m,\ 80\ m]$ and $[-40\ m,\ 40\ m]$ on the $x$ and $y$ axes respectively. The second uses $0.16\ m$ pillars with a grid area of $[0\ m,\ 81.92\ m]$ and $[-40.96\ m,\ 40.96\ m]$ on the $x$ and $y$ axes respectively. This difference in the grid size is due to the downsampling and upsampling of the tensors through the neck and backbone.

We use the OpenPCDet library~\cite{openpcdet2020} for PillarNet as well as for the rulebook-based sparse backbone in Aerial-PointPillars. All implementations use a z-axis of $[-10\ m, 10\ m]$. For performance testing, a dataset containing $1,603$ frames of drones from the Newcastle scenario, using $1,282$ for training and $321$ for testing. All models were trained for 128 epochs on a Nvidia RTX 3080 Ti GPU with an IoU threshold of 0.3.

\subsection{PandaSet Testing}\label{sec:pandset_method}

To the best of our knowledge, there are no other drone-based datasets against which to test SSCATeR, and only one dataset within the domain of autonomous driving-based object detection which includes per-point timestamps. As such, we test SSCATeR on this dataset by using an $8\ s$ sequence from PandaSet~\cite{pandaset}, the maximum duration of their sequences. The test will measure the runtime of a SSCATeR PointPillars backbone against a scatter-based PointPillars backbone.

The PandaSet data has a much higher number of points than the drone-based datasets, in part due to the returns from the road surface, and also as it is ground-based and therefore does not need to remain within the same strict SWaP parameters as a drone. We use the data from the front facing LiDAR, a PandarGT with the equivalent to 150 channels. To avoid random sampling in this sequence, the maximum number of pillars must be increased to $13,260$, whilst the maximum number of points per pillar must be increased to $1,098$. The sequence is summarized in Table \ref{pandaset_summary_table}, where the 801 frames refer to $100\ ms$ frames that increment in $10\ ms$ time intervals.

\begin{table}
\caption{Summary of PandaSet sequence data.} 
\label{pandaset_summary_table}
\setlength{\tabcolsep}{3pt}
\begin{tabular}{|p{35pt}|p{30pt}|p{50pt}|p{45pt}|p{60pt}|}
\hline
Dataset&Total Frames&Average Points Per Frame&Points Range Per Frame&Environment and Scenario\\
\hline
PandaSet&$801$&$61019.23$&$[1038, 65844]$&Ground-based vehicle driving in a city.\\
\hline
\end{tabular}
\end{table}

\subsection{LiDAR-to-LiDAR Crosstalk Interference Testing}\label{sec:crosstalk}

Given the use case of multi-drone swarms, we also explore the effects of crosstalk interference between two drone-mounted LiDAR units on the accuracy of the network. To test this, we have collected a dataset at Strathaven Airfield, using two L1 LiDAR systems mounted on two DJI M300 drones during a survey of an aircraft hangar. We have included scenarios of both indirect crosstalk and direct crosstalk. For the indirect crosstalk, where pulses from one LiDAR may reflect off a surface and erroneously be captured by another LiDAR, we stationed the drones at $45\degree$ and $-45\degree$ angles facing toward the hangar, as well as having the surveying drone move directly in front of the monitoring drone. For the direct crosstalk, where pulses from one LiDAR are fired towards and enter another LiDAR system directly, the two drones faced one another with their LiDARs facing forward. We utilize $3,635$ frames for training, and $1,810$ frames for testing. We train both models over $128$ epochs using the same hyperparameters as those described in section \ref{sec:sparse-backbone}.

\section{Results}
\label{sec:results}

\subsection{Analysis and Evaluation of Pillar Feature Net Modifications}

\begin{table}
\caption{Comparison of SSCATeR and Sparse Convolution Mean Runtimes on AGX XAVIER} 
\label{table1}
\setlength{\tabcolsep}{3pt}
\begin{tabular}{|p{35pt}|p{35pt}|p{50pt}|p{65pt}|p{35pt}|}
\hline
Image \par Channels& 
Number of \par Filters& 
SSCATeR Runtime ($ms$)&
Sparse Convolution \par Runtime ($ms$)&
Difference ($\%$) \\
\hline
$16$& $8$&    $0.333142$& $0.991550$& $66.40$ \\
$16$& $16$&   $0.443594$& $1.566631$& $71.68$ \\
$16$& $32$&   $0.699696$& $2.823243$& $75.22$ \\
$16$& $64$&   $1.216464$& $5.370997$& $77.35$ \\
$16$& $128$&  $2.235108$& $10.341904$& $78.39$ \\
\hline
$32$& $8$&    $0.345238$& $1.054077$& $67.25$ \\
$32$& $16$&   $0.463043$& $1.657492$& $72.06$ \\
$32$& $32$&   $0.735343$& $2.979420$& $75.32$ \\
$32$& $64$&   $1.274623$& $5.657116$& $77.47$ \\
$32$& $128$&  $2.405750$& $11.049710$& $78.23$ \\
\hline
$64$& $8$&    $0.403079$& $1.211235$& $66.72$ \\
$64$& $16$&   $0.536994$& $1.937511$& $72.28$ \\
$64$& $32$&   $0.877795$& $4.061695$& $78.39$ \\
$64$& $64$&   $1.547171$& $7.212030$& $78.55$ \\
$64$& $128$&  $2.956234$& $13.873355$& $78.69$ \\
\hline
$128$& $8$&   $0.507775$& $2.547465$& $80.07$ \\
$128$& $16$&  $0.737575$& $3.446114$& $78.60$ \\
$128$& $32$&  $1.169908$& $5.015102$& $76.67$ \\
$128$& $64$&  $1.934333$& $8.453566$& $77.12$ \\
$128$& $128$& $3.557213$& $15.333513$& $76.80$ \\
\hline
\end{tabular}
\label{runtime_table_xavier}
\end{table}

This section details the results of our modifications made to the PFN stage of the network. Through the use of change maps, the number of active sites that require convolution computation is reduced by $72.8\%$. The modifications also mean that, as a result of reusing data from previous passes through the network, we find that the number operations in our PFN stage is reduced by $70.97\%$.

The Livox Avia and L1 LiDARs are capable of generating $24,000$ points every $100\ ms$. However, due to the forward looking mode of operation, a significant number of points are not reflected as they do not make contact with a surface. In Fig. \ref{fig:nc_points_and_pillars} the point cloud frame contains $6,437$ points captured over $100\ ms$.

The pillar grids are comprised of an $80.64\ m \times 80.64\ m$ detection area discretized into $504 \times 504$ grid of $0.16\ m$ pillars in the horizontal dimensions, resulting in $254,016$ pillars in total. The same pillar grid is produced by the one-shot and striding point-binning methods. The pillar grids produced by each of the $10\ ms$ strides are shown in Fig. \ref{fig:nc_pillars}. 

On average, changed sites make up only 27.2\% of active sites per 10 ms sliding window frame, meaning an average reduction of 72.8\% in required convolutions. The change maps for each stride are shown in Fig. \ref{fig:nc_change}. Across $10,000$ $10\ ms$ frames from the Newcastle dataset, we measured that the number of active sites ranged between $1,552$ and $2,533$, whereas changed sites ranged between $169$ and $1,154$. Across all frames, the mean active sites was $1,781.12$, and mean changed sites was $482.78$.

\begin{figure}[htb]
    \includegraphics[width=\columnwidth]{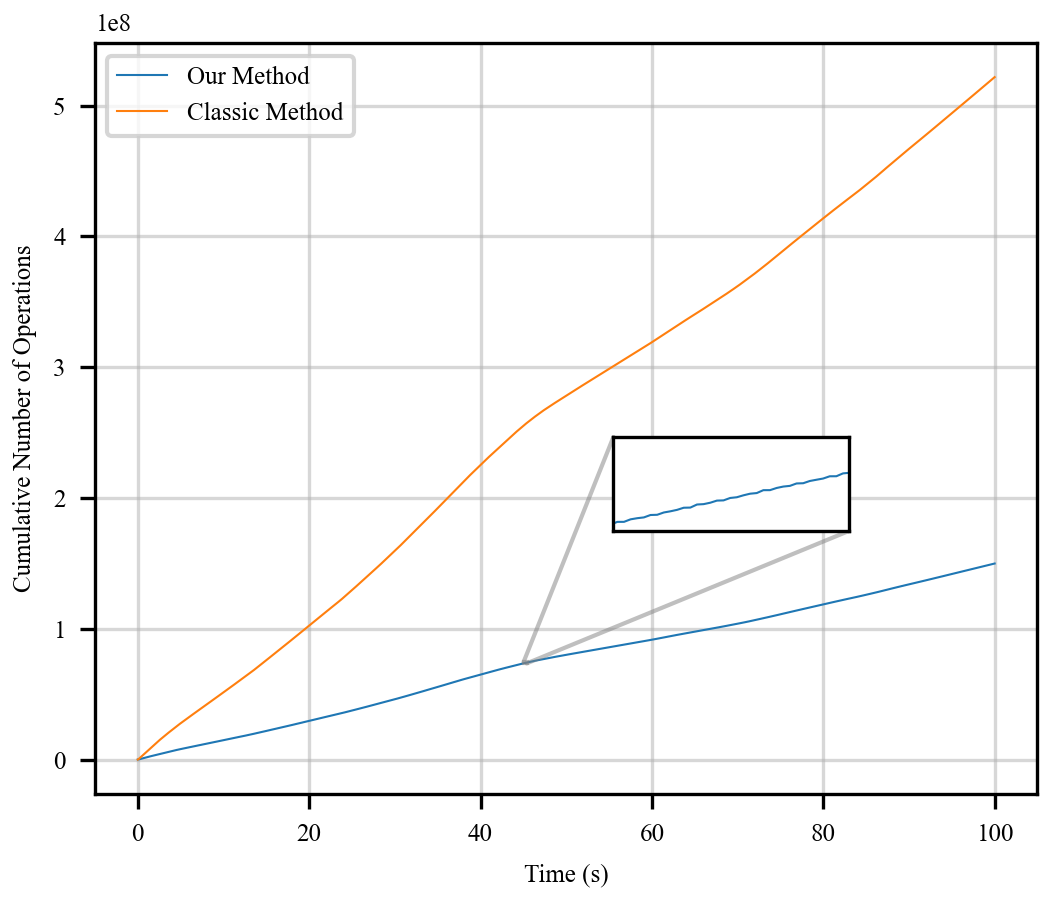}
    \caption{Comparison of cumulative counts of operations on the Newcastle dataset over $100\ s$, comparing our method and the standard method.}
    \label{fig:pillar_op_count}
\end{figure}

Fig. \ref{fig:nc_change} shows the subset of sites that need to be operated on in each $10\ ms$ frame update. Fig. \ref{fig:pillar_op_count} shows the corresponding count of operations required to generate the pillar feature maps, comparing SSCATeR and the original PointPillars methods applied to a $100$ second sequence of frames. On average, the SSCATeR PFN stage requires $70.97\%$ fewer operations per frame. We have benchmarked the algorithm at $0.2\ ms$ on the Nvidia AGX Orin ($32\ GB$) detailed below.

\begin{table}
\caption{Comparison of SSCATeR and Sparse Convolution Mean Runtimes on AGX Orin} 
\label{table2}
\setlength{\tabcolsep}{3pt}
\begin{tabular}{|p{35pt}|p{35pt}|p{50pt}|p{65pt}|p{35pt}|}
\hline
Image \par Channels& 
Number of \par Filters& 
SSCATeR Runtime ($ms$)&
Sparse Convolution \par Runtime ($ms$)&
Difference ($\%$) \\
\hline
$16$& $8$&    $0.286118$& $0.693747$& $58.76$ \\
$16$& $16$&   $0.236026$& $0.700770$& $66.32$ \\
$16$& $32$&   $0.335697$& $1.137218$& $70.48$ \\
$16$& $64$&   $0.564989$& $2.067727$& $72.68$ \\
$16$& $128$&  $1.050329$& $4.062497$& $74.15$ \\
\hline
$32$& $8$&    $0.285290$& $0.659575$& $56.75$ \\
$32$& $16$&   $0.277278$& $0.732610$& $62.15$ \\
$32$& $32$&   $0.415773$& $1.200764$& $65.37$ \\
$32$& $64$&   $0.684382$& $2.170695$& $68.47$ \\
$32$& $128$&  $1.156805$& $4.221957$& $72.60$ \\
\hline
$64$& $8$&    $0.267731$& $0.669489$& $60.01$ \\
$64$& $16$&   $0.326457$& $0.819357$& $60.16$ \\
$64$& $32$&   $0.465815$& $2.491423$& $81.30$ \\
$64$& $64$&   $0.735168$& $3.430603$& $78.57$ \\
$64$& $128$&  $1.231972$& $5.627393$& $78.11$ \\
\hline
$128$& $8$&   $0.345320$& $2.283725$& $84.88$ \\
$128$& $16$&  $0.461963$& $2.531632$& $81.75$ \\
$128$& $32$&  $0.624681$& $3.074259$& $79.68$ \\
$128$& $64$&  $0.903414$& $4.237840$& $78.68$ \\
$128$& $128$& $1.486286$& $6.407384$& $76.80$ \\
\hline
\end{tabular}
\label{runtime_table_orin}
\end{table}

\subsection{Backbone Timing Analysis} \label{analysis}

The convolutions were timed over $100\ s$ across two methods. In doing so, we find that SSCATeR is on average $2.31$ to $6.61$ times faster than scatter-based sparse convolution across a single convolutional layer, and on average is $3.83$ to $4.13$ times faster across the entire backbone.

To test the single convolution, a single convolutional layer was timed to compare scatter-based sparse convolutions and SSCATeR. The single convolutional layer used a $3 \times 3$ single stride kernel, with $16,\ 32,\ 64,$ or $128$ input channels, and $8,\ 16,\ 32,\ 64,$ or $128$ filters. The input used the $504 \times 504$-site feature map produced by the pillar feature net. We then tested the full backbone, consisting of the three blocks, upsampling, and concatenation described in the methodology and shown in Fig. \ref{fig:pp_arch}. The input for the full backbone was also the $504 \times 504$-site feature map produced by the pillar feature net.

Both the single layer and full backbone were compared across $10,000$ frames, totalling $100\ s$ of data. The Nvidia AGX Xavier ($32\ GB$) and Orin ($32\ GB$) series were used for testing, with Jetpack 5.1.3, CUDA 11.4 and cuDNN 8.6 installed. The Xavier has 512 Volta CUDA cores, while the Orin has 2048 Ampere CUDA cores.

\begin{table}
\caption{Comparison of SSCATeR and Sparse Backbone Mean Runtimes} 
\label{table3}
\setlength{\tabcolsep}{3pt}
\begin{tabular}{|p{22.5pt}|p{47.5pt}|p{47.5pt}|p{65pt}|p{37.5pt}|}
\hline
Board & 
Block(s) & 
SSCATeR\par Runtime ($ms$) & 
Sparse Convolution\par Runtime ($ms$) & 
Difference ($\%$) \\
\hline
\multirow{4}{*}{Xavier} & Block 1       & $2.494039$    & $13.191079$   & $81.09$       \\
                        & Block 2       & $4.406001$    & $16.385524$   & $73.11$       \\
                        & Block 3       & $5.901391$    & $20.112170$   & $70.66$       \\
                        & All Blocks    & $12.801430$   & $49.688773$   & $74.24$       \\
\hline
\multirow{4}{*}{Orin}   & Block 1       & $1.691551$    & $6.750310$    & $74.94$        \\
                        & Block 2       & $2.752337$    & $10.735333$   & $74.36$        \\
                        & Block 3       & $3.550490$    & $10.335987$   & $65.65$        \\
                        & All Blocks    & $7.994378$    & $27.821631$   & $71.27$       \\
\hline
\end{tabular}
\label{backboneComparison}
\end{table}

\begin{figure}[!h]
    \includegraphics[width=\columnwidth]{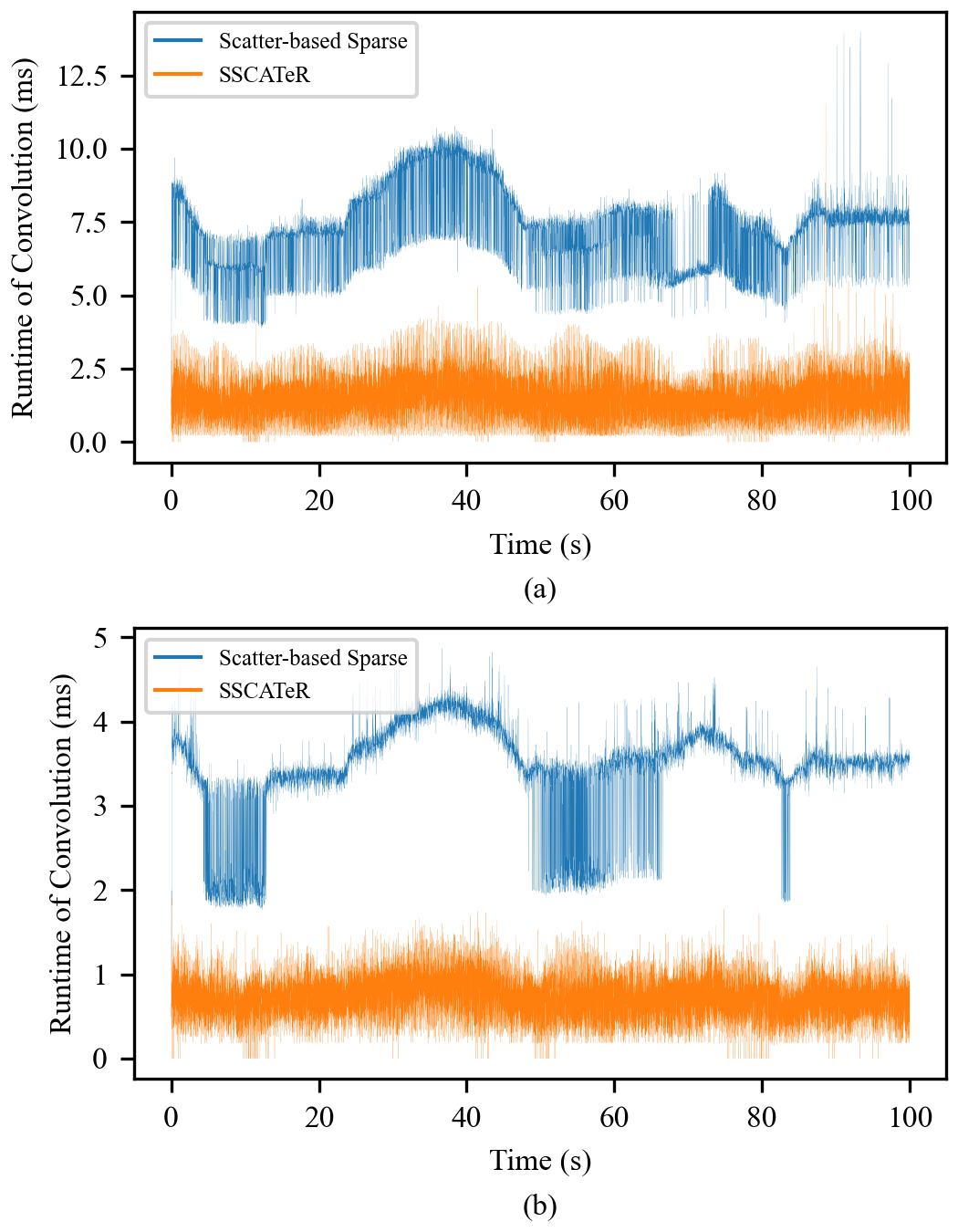}
\caption{Run time (milliseconds) of SSCATeR convolution on the Newcastle dataset, compared with scatter-based sparse convolution on the AGX Xavier (a) and AGX Orin (b); Size of image = $504 \times 504$, filter size = $3 \times 3$.}
\label{fig:performanceComparison}
\end{figure}

\begin{figure}[!h]
    \includegraphics[width=\columnwidth]{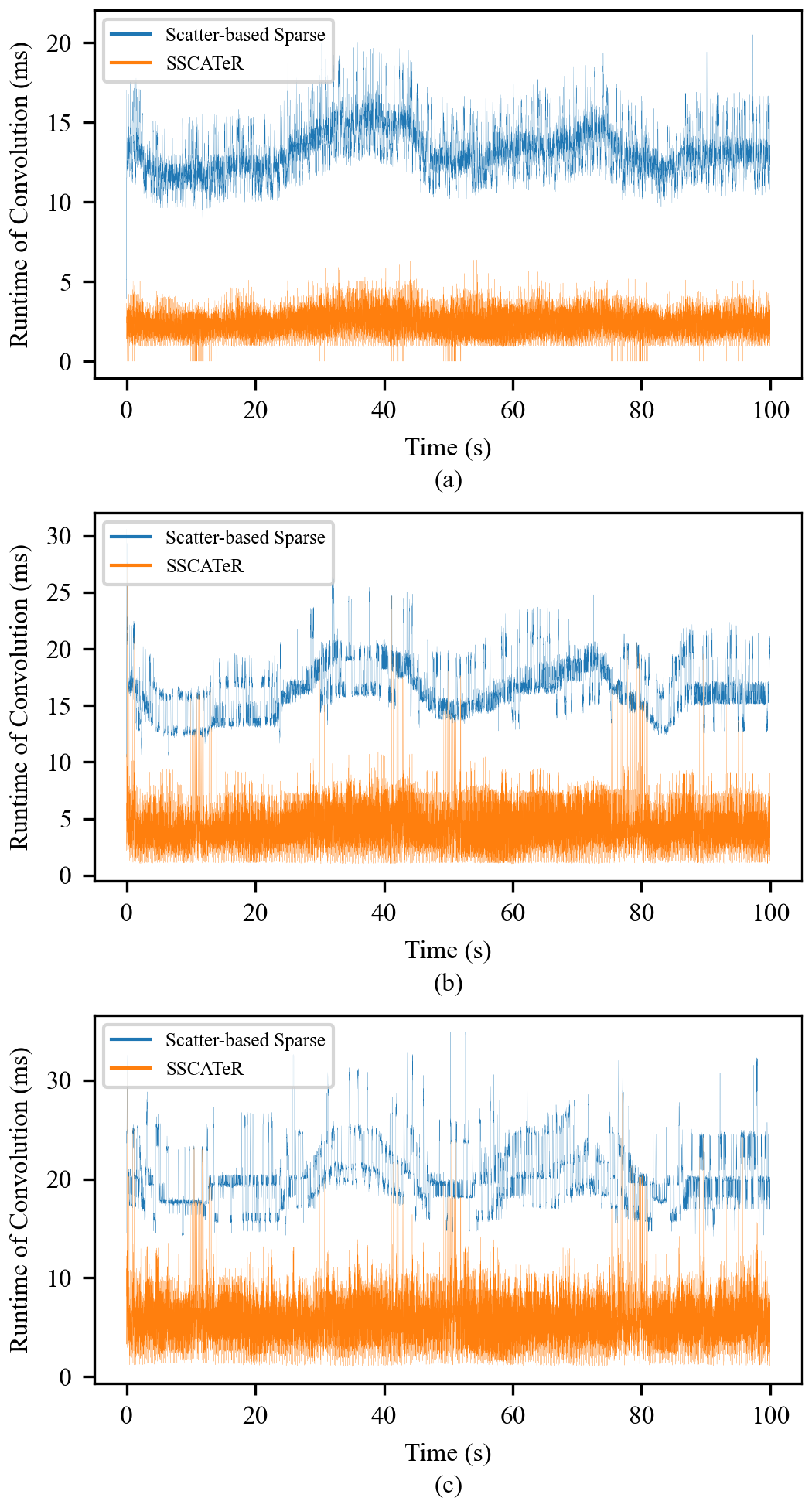}
\caption{Average run time (milliseconds) of SSCATeR on Newcastle dataset compared with scatter-based sparse for each convolutional block (a, b, c) in the backbone on AGX Xavier.}
\label{fig:xavierblockComparison}
\end{figure}

\begin{figure}[!h]
    \includegraphics[width=\columnwidth]{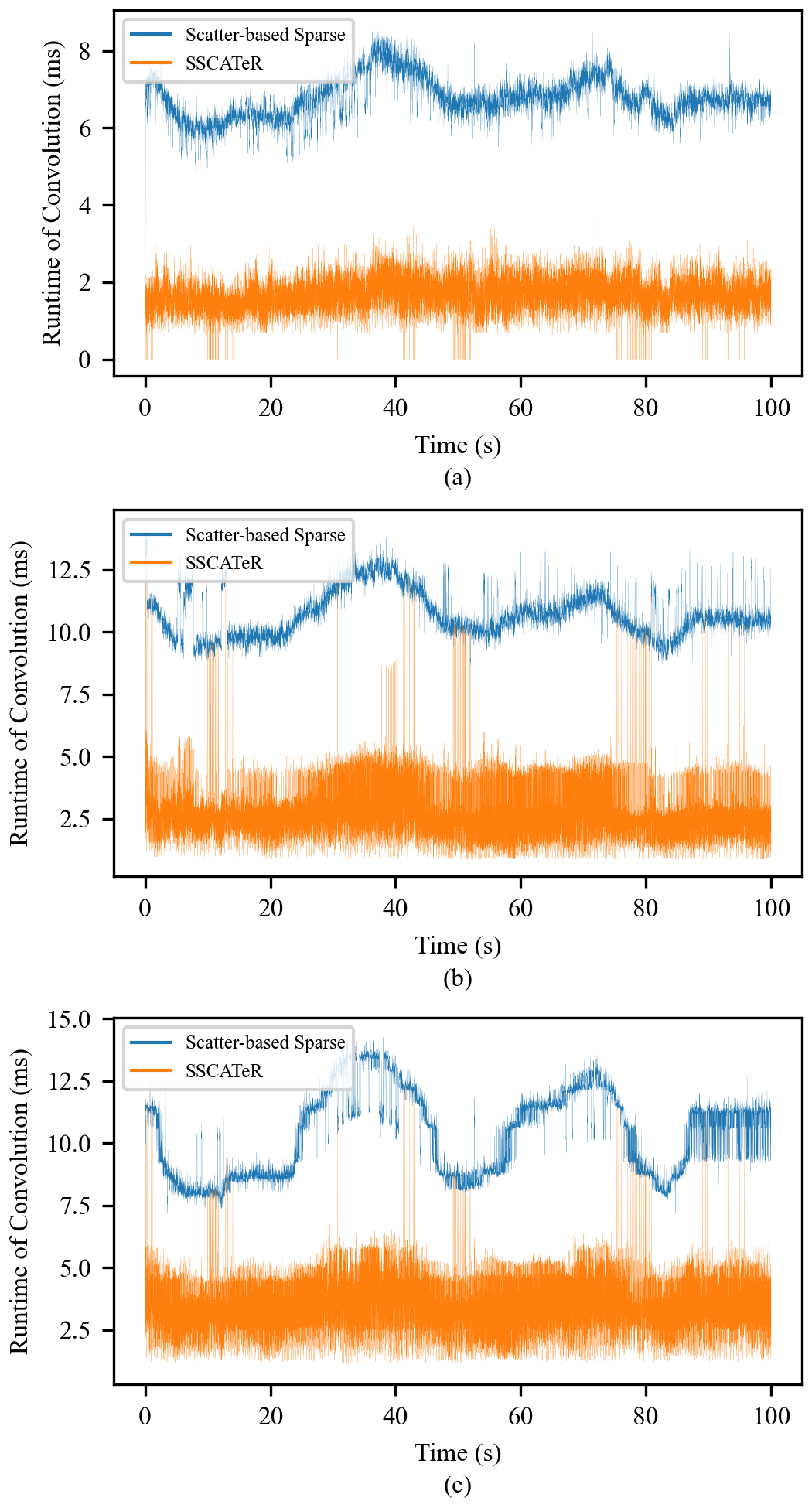}
\caption{Average run time (milliseconds) of SSCATeR on Newcastle dataset compared with scatter-based sparse for each convolutional block (a, b, c) in the backbone on AGX Orin.}
\label{fig:orinblockComparison}
\end{figure}

Both systems were running in "MaxN" power mode. For the Xavier's CPU, this means eight active CPU cores with a max frequency of $2265.6\ MHz$. For the Xavier's GPU, this means 4 texture processing clusters (TPC) and a max GPU frequency of $1377\ MHz$, with 2 deep learning accelerator (DLA) cores at a max frequency of $1395.2\ MHz$. For the Orin's CPU, this means eight active CPU cores with a max frequency of $2188.8\ MHz$. For the Orin's GPU, this means 7 TPC and a $930.75\ MHz$ max frequency, with 2 DLA cores with a max frequency of $1408\ MHz$.

Table \ref{runtime_table_xavier} and Table \ref{runtime_table_orin} show the mean runtime results of the single convolutional layer comparison. Table \ref{runtime_table_xavier} benchmarks on the Xavier, and it can be seen that SSCATeR is between $2.98$ and $5.02$ times faster than scatter-based sparse convolution, with a mean of $4.13$. Table \ref{runtime_table_orin} benchmarks on the Orin, and shows that SSCATeR is between $2.31$ and $6.61$ times faster than scatter-based sparse convolution, with a mean of $3.83$.

Fig. \ref{fig:performanceComparison} display the runtime of the $64$ input channel and convolutional filters from Tables \ref{runtime_table_xavier} and \ref{runtime_table_orin} over $100\ s$ of consecutive frames. This specific combination of values were chosen as an example here as they are the first convolutional values in the PointPillars backbone. The graphs show that SSCATeR is consistently faster than scatter-based sparse convolutions.

In all of the convolution comparisons over time, spikes above the baseline can be seen. These spikes are caused by a combination of an increase in the number of active sites, data caching, memory access allocation and background processes occurring in the Xavier and Orin.

The full backbone testing follows the input and output pipeline described in the methodology, block 1 passes it's data to block 2 before upsampling, and the same occurs from block 2 to block 3. It can be seen in Fig. \ref{fig:xavierblockComparison} and Fig. \ref{fig:orinblockComparison} that the runtime differences between scatter-based sparse and SSCATeR become more pronounced in the lower blocks. Whilst block 1 has the largest input image size of the three, it also has the least amount of convolutional layers with four, compared to six in blocks 2 and 3. Each block also consists of more output channels than the previous layer, which further increases the computation. Table \ref{backboneComparison} displays the mean runtimes of each block, as well as the full backbone, on both the Xavier and Orin. Overall, the SSCATeR backbone is $3.88$ times faster on the Xavier, and $3.48$ times faster on the Orin, than the scatter-based sparse approach. It is also noted that whilst the runtime difference in $ms$ increases in the lower blocks, the percentage difference decreases.

As the incoming data is processed in $10\ ms$ frames, it is important that the backbone can process the data faster than this. Across the $10,000$ individual backbone runtimes, scatter-based sparse was slower than $10\ ms$ $100.00\%$ of the time on the Xavier and $99.99\%$ of the time on the Orin. SSCATeR was slower than $10\ ms$ $67.59\%$ of the time on the Xavier, and $20.63\%$ of the time on the Orin. It is worth noting that SSCATeR on the Orin was able to return to real-time, i.e. less than 10ms processing time, each time it ran over, recording only $25.29\%$ above $10\ ms$. Of the time spent above $10\ ms$ the mean processing latency was just $12.977961\ ms$. The longest processing latency was $37.596770\ ms$, still significantly faster than waiting for $100\ ms$ of data to be collected. Overall, the fastest full backbone time that SSCATeR recorded was $1.731936\ ms$ on the Orin, in comparison to the fastest scatter-based sparse time of $5.919292\ ms$, also on the Orin.

\begin{table}
\caption{Comparison of SSCATeR and Sparse Convolution Mean Runtimes on AGX XAVIER} 
\label{table4}
\setlength{\tabcolsep}{3pt}
\begin{tabular}{|p{35pt}|p{35pt}|p{50pt}|p{65pt}|p{35pt}|}
\hline
Image \par Channels& 
Number of \par Filters& 
SSCATeR Runtime ($ms$)&
Sparse Convolution \par Runtime ($ms$)&
Difference ($\%$) \\
\hline
$16$& $8$&    $0.423965$& $0.615609$& $31.13$ \\
$16$& $16$&   $0.467563$& $0.752498$& $37.87$ \\
$16$& $32$&   $0.568126$& $1.063325$& $46.57$ \\
$16$& $64$&   $0.759285$& $1.701853$& $55.38$ \\
$16$& $128$&  $1.120866$& $2.965487$& $62.20$ \\
\hline
$32$& $8$&    $0.433403$& $0.646064$& $32.92$ \\
$32$& $16$&   $0.491483$& $0.808127$& $39.18$ \\
$32$& $32$&   $0.588914$& $1.132326$& $47.99$ \\
$32$& $64$&   $0.795996$& $1.814832$& $56.14$ \\
$32$& $128$&  $1.183320$& $3.157302$& $62.52$ \\
\hline
$64$& $8$&    $0.467520$& $0.710288$& $34.18$ \\
$64$& $16$&   $0.527412$& $0.881999$& $40.20$ \\
$64$& $32$&   $0.648553$& $1.285879$& $49.56$ \\
$64$& $64$&   $0.882777$& $2.074164$& $57.44$ \\
$64$& $128$&  $1.353221$& $3.689370$& $63.32$ \\
\hline
$128$& $8$&   $0.512785$& $0.917168$& $44.09$ \\
$128$& $16$&  $0.588400$& $1.137185$& $48.26$ \\
$128$& $32$&  $0.739758$& $1.658357$& $55.39$ \\
$128$& $64$&  $1.041443$& $2.651863$& $60.73$ \\
$128$& $128$& $1.652243$& $4.750220$& $65.22$ \\
\hline
\end{tabular}
\label{runtime_table_xavier_hb}
\end{table}

\begin{table}
\caption{Comparison of SSCATeR and Sparse Convolution Mean Runtimes on AGX Orin} 
\label{table5}
\setlength{\tabcolsep}{3pt}
\begin{tabular}{|p{35pt}|p{35pt}|p{50pt}|p{65pt}|p{35pt}|}
\hline
Image \par Channels& 
Number of \par Filters& 
SSCATeR Runtime ($ms$)&
Sparse Convolution \par Runtime ($ms$)&
Difference ($\%$) \\
\hline
$16$& $8$&    $0.208015$& $0.281946$& $26.22$ \\
$16$& $16$&   $0.198560$& $0.313397$& $36.64$ \\
$16$& $32$&   $0.258801$& $0.438541$& $40.99$ \\
$16$& $64$&   $0.346100$& $0.727263$& $52.41$ \\
$16$& $128$&  $0.519415$& $1.296245$& $59.93$ \\
\hline
$32$& $8$&    $0.307917$& $0.418566$& $26.44$ \\
$32$& $16$&   $0.203350$& $0.353256$& $42.44$ \\
$32$& $32$&   $0.276265$& $0.503081$& $45.09$ \\
$32$& $64$&   $0.384159$& $0.783178$& $50.95$ \\
$32$& $128$&  $0.572222$& $1.381012$& $58.57$ \\
\hline
$64$& $8$&    $0.227647$& $0.329089$& $30.83$ \\
$64$& $16$&   $0.271178$& $0.423783$& $36.01$ \\
$64$& $32$&   $0.327378$& $0.551022$& $40.59$ \\
$64$& $64$&   $0.428888$& $0.848506$& $49.45$ \\
$64$& $128$&  $0.617013$& $1.472116$& $58.09$ \\
\hline
$128$& $8$&   $0.284552$& $0.445916$& $36.19$ \\
$128$& $16$&  $0.315660$& $0.478349$& $34.01$ \\
$128$& $32$&  $0.382566$& $0.651080$& $41.24$ \\
$128$& $64$&  $0.480451$& $0.979943$& $50.97$ \\
$128$& $128$& $0.718721$& $1.728919$& $58.43$ \\
\hline
\end{tabular}
\label{runtime_table_orin_hb}
\end{table}

\begin{table}
\caption{Comparison of SSCATeR and Sparse Backbone Mean Runtimes} 
\label{table6}
\setlength{\tabcolsep}{3pt}
\begin{tabular}{|p{22.5pt}|p{47.5pt}|p{47.5pt}|p{65pt}|p{37.5pt}|}
\hline
Board & 
Block(s) & 
SSCATeR\par Runtime ($ms$) & 
Sparse Convolution\par Runtime ($ms$) & 
Difference ($\%$) \\
\hline
\multirow{4}{*}{Xavier} & Block 1       & $1.509864$    & $2.749642$    & $45.09$        \\
                        & Block 2       & $3.226693$    & $6.972936$    & $53.73$        \\
                        & Block 3       & $6.229286$    & $12.921595$   & $51.79$        \\
                        & All Blocks    & $10.965843$   & $22.644173$   & $51.57$       \\
\hline
\multirow{4}{*}{Orin}   & Block 1       & $1.166546$    & $1.798241$    & $35.13$        \\
                        & Block 2       & $2.101325$    & $3.600370$    & $41.64$        \\
                        & Block 3       & $3.201389$    & $5.817380$    & $44.97$        \\
                        & All Blocks    & $6.469260$    & $11.215991$   & $42.32$        \\
\hline
\end{tabular}
\label{backboneComparison_hb}
\end{table}

\subsection{Backbone Testing on Hamburg Dataset}

As the backbone is the most computationally expensive component of the network's architecture, further performance analysis was conducted using a different scenario. The Hamburg dataset, described in \ref{sec:datasets}, is used for this purpose.

\begin{figure}[!h]
    \includegraphics[width=\columnwidth]{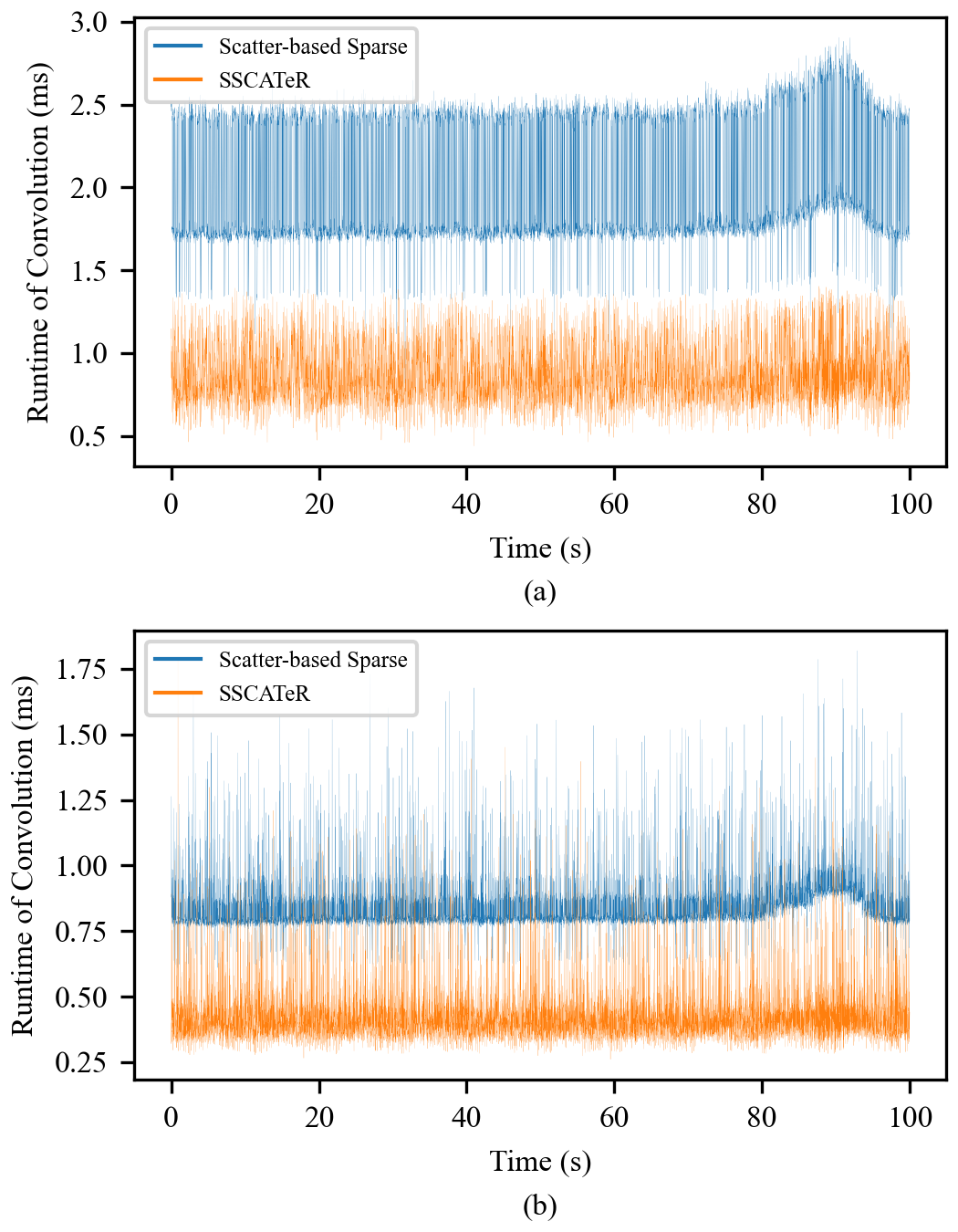}
\caption{Run time (milliseconds) of SSCATeR convolution on the Hamburg dataset, compared with scatter-based sparse convolution on the AGX Xavier (a) and AGX Orin (b); Size of image = $504 \times 504$, filter size = $3 \times 3$.}
\label{fig:performanceComparison_hb}
\end{figure}

\begin{figure}[!h]
    \includegraphics[width=\columnwidth]{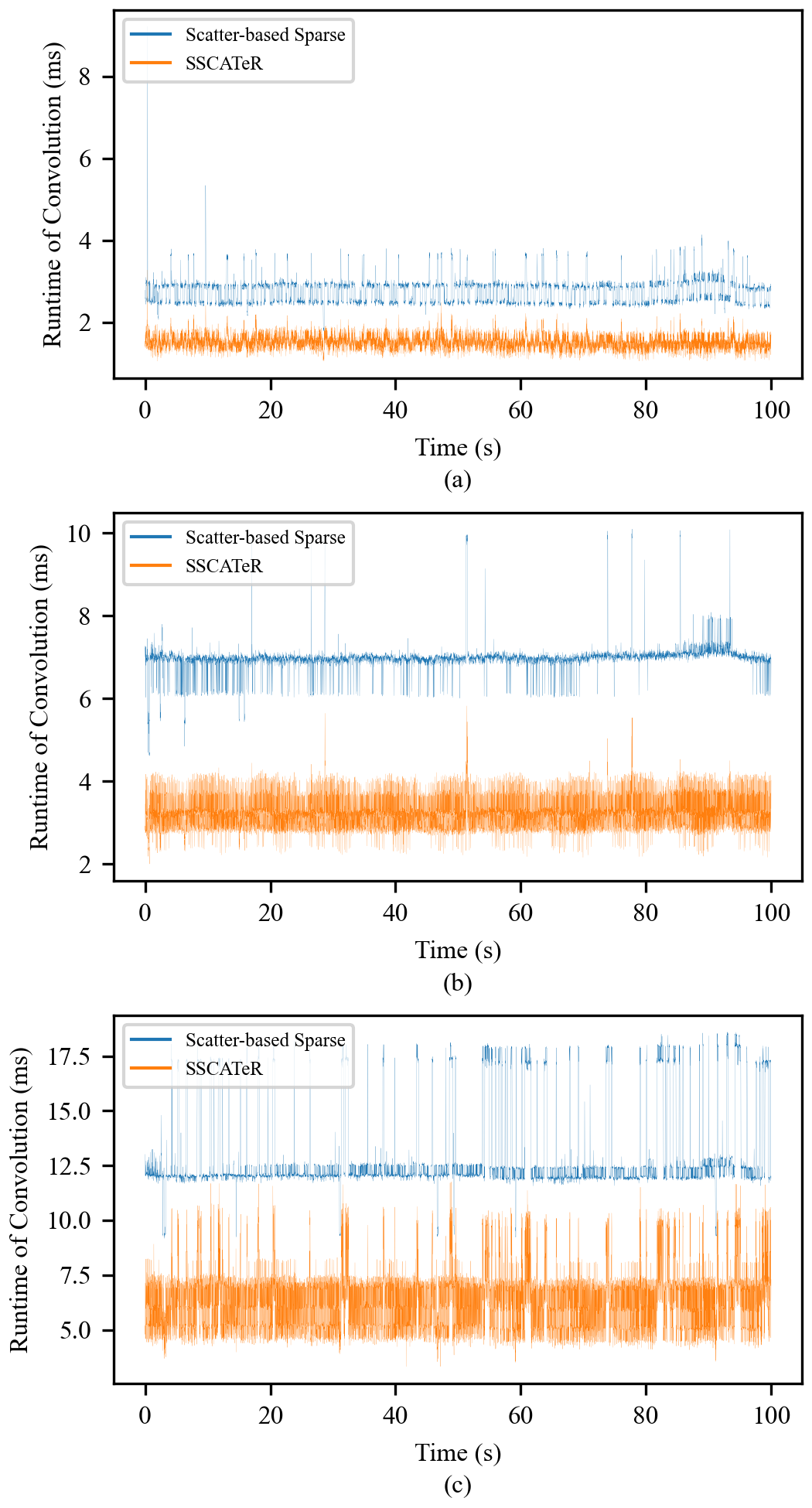}
\caption{Average run time (milliseconds) of SSCATeR on Hamburg dataset compared with scatter-based sparse for each convolutional block (a, b, c) in the backbone on AGX Xavier.}
\label{fig:xavierblockComparison_hb}
\end{figure}

\begin{figure}[!h]
    \includegraphics[width=\columnwidth]{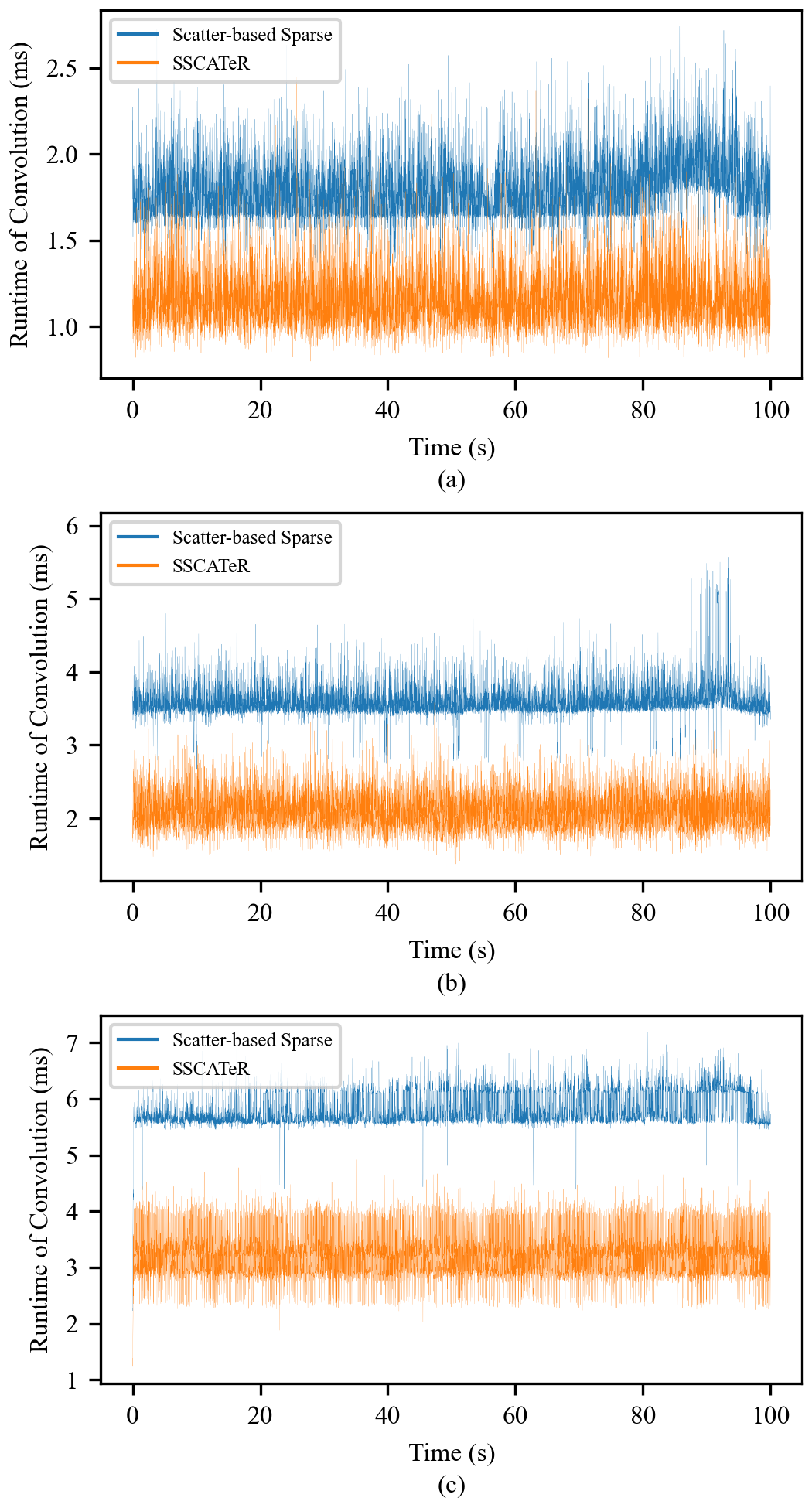}
\caption{Average run time (milliseconds) of SSCATeR on Hamburg dataset compared with scatter-based sparse for each convolutional block (a, b, c) in the backbone on AGX Orin.}
\label{fig:orinblockComparison_hb}
\end{figure}

In the Xavier benchmarks for the single convolutional layer test in Table \ref{runtime_table_xavier_hb} it can be seen that SSCATeR is between $1.45$ and $2.88$ times faster than scatter-based sparse convolution, with a mean of $2.07$. On the Orin benchmarks of the same test in Table \ref{runtime_table_orin_hb} SSCATeR is between $1.35$ and $2.50$ times faster than scatter-based sparse convolution, with a mean of $1.84$. The runtime of the $64$ input channel and convolutional filters from Tables \ref{runtime_table_xavier_hb} and \ref{runtime_table_orin_hb} over $100\ s$ of consecutive frames is displayed in Fig. \ref{fig:performanceComparison_hb}. These graphs show that, in terms of the Hamburg dataset, SSCATeR is consistently around twice as fast as the scatter-based sparse convolutions.

The full backbone testing for the Hamburg dataset in Fig. \ref{fig:xavierblockComparison_hb} and Fig. \ref{fig:orinblockComparison_hb} correlate with the results from the Newcastle dataset, in that the runtime differences become more pronounced in the lower blocks. The mean runtimes of each block and the full backbone for both hardware systems is shown in Table \ref{backboneComparison_hb}. On both hardware systems in the Hamburg dataset the performance gains were lower than on the Newcastle dataset, due to the lower number of active and changed sites on average. Here, the SSCATeR backbone is $2.06$ and $1.73$ times faster than scatter-based sparse convolution on the Xavier and Orin respectively.

When comparing the ability to perform in real-time, the results were similar in both datasets, however in Hamburg, SSCATeR always completed in under $10\ ms$ on the Orin. Scatter-based sparse was slower than $10\ ms$ $100.00\%$ of the time on the Xavier and $99.71\%$ of the time on the Orin. SSCATeR on the Xavier was slower than $10\ ms$ $63.77\%$ of the time. SSCATeR's fastest full backbone time was $3.19875\ ms$ on the Orin, and the fastest scatter-based sparse time of $5.95254\ ms$, also on the Orin.

\subsection{Comparison of Different Architectures}

We further compare the backbone latency of different architectures by testing them on the Newcastle dataset, we also compare the performance of the architectures on a drone-based object detection dataset.

\begin{figure}[thb]
    \includegraphics[width=\columnwidth]{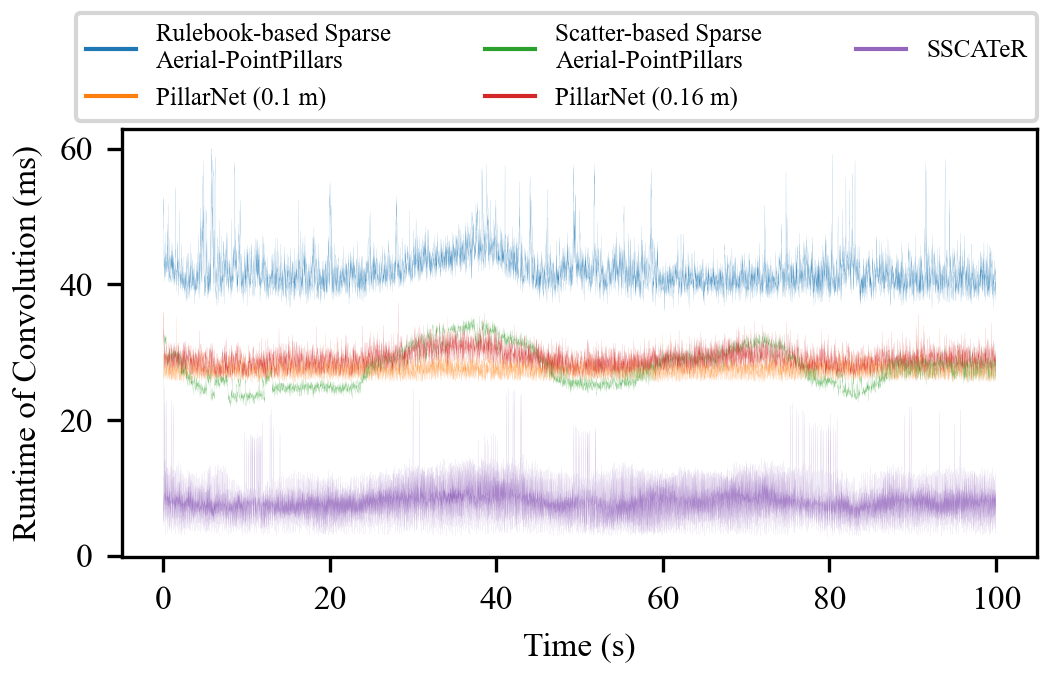}
    \caption{Backbone runtime comparison of the Aerial-PointPillars, PillarNet, and SSCATeR implementations, measured in milliseconds. PillarNet timings include both the backbone and neck of the architecture.}
    \label{fig:architecure_latency_comparison}
\end{figure}
\begin{table}[]
\centering
\caption{Comparison of Mean Runtime Across Architectures}
\label{tab:arch_comparison_latency}
\begin{tabular}{|p{74pt}|p{47pt}|p{42pt}|p{42pt}|}
\hline
Architecture                                                                        & Mean Runtime ($ms$) & Max Runtime ($ms$) & Min Runtime ($ms$) \\
\hline
PillarNet ($0.1\ m$)                                                                & $27.566420$         & $35.671234$        & $25.338888$        \\
\hline
PillarNet ($0.16\ m$)                                                               & $29.113901$         & $37.249804$        & $25.866270$        \\
\hline
\begin{tabular}[c]{@{}l@{}}Rulebook-Based Sparse\\ Aerial-PointPillars\end{tabular} & $42.066090$         & $60.034037$        & $36.126137$        \\
\hline
\begin{tabular}[c]{@{}l@{}}Scatter-Based Sparse\\ Aerial-PointPillars\end{tabular}  & $27.821631$         & $35.793200$        & $6.746670$         \\
\hline
SSCATeR (Ours)                                                                             & $7.994378$          & $24.838100$        & $2.742292$         \\
\hline
\end{tabular}
\end{table}

The runtime of each architecture over the $10,000$ frames from the Newcastle dataset can be seen in Fig. \ref{fig:architecure_latency_comparison}. The comparisons between SSCATeR and scatter-based sparse Aerial-PointPillars have been documented previously, but it can be seen that SSCATeR remains the fastest. Table \ref{tab:arch_comparison_latency} shows the mean, max, and min runtimes across the architectures, and it can be seen that SSCATeR is the only architecture with a mean runtime in the sub-$10\ ms$ real-time processing range. In this scenario, SSCATeR achieves a $5.26$-times improvement over the rulebook-based sparse Aerial-PointPillars backbone, which is an $81.00\%$ reduction in processing time. Compared to the PillarNet architectures, which are also using rulebook-based sparse convolutions, the latency reduction is similar to the reduction achieved over the scatter-based sparse Aerial-PointPillars backbone.

\begin{table}[]
\centering
\caption{Performance Metrics for PillarNet and Aerial-PointPillars Models}
\label{tab:architecture_comparison_performance}
\begin{tabular}{|p{65pt}|p{30pt}|p{25pt}|p{30pt}|p{40pt}|}
\hline
Architecture          & Precision & Recall & F1-Score & mAP   \\
\hline
PillarNet ($0.1\ m$)  & $96.552$    & $95.950$ & $96.250$   & $95.661$ \\
\hline
PillarNet ($0.16\ m$) & $97.508$    & $97.508$ & $97.508$   & $97.138$ \\
\hline
Aerial-PointPillars   & $79.189$    & $91.277$ & $84.805$   & $89.738$ \\
\hline
\end{tabular}
\end{table}

The results of the performance testing can be seen in Table \ref{tab:architecture_comparison_performance}, where the best PillarNet model achieves an mAP of $97.138\%$, a $7.4\%$ increase on the Aerial-PointPillars mAP. This difference is expected, as the literature documents PillarNet as a higher performance model than PointPillars. From the perspective of a SAD use-case, the recall is more important than the precision, as missing a true positive detection can result in a safety-critical accident. With this in mind, Aerial-PointPillars is closer to the best PillarNet model in terms of recall, scoring only $6.23\%$ less.

\subsection{Testing on PandaSet}

\begin{figure}[bth]
    \includegraphics[width=3.49in]{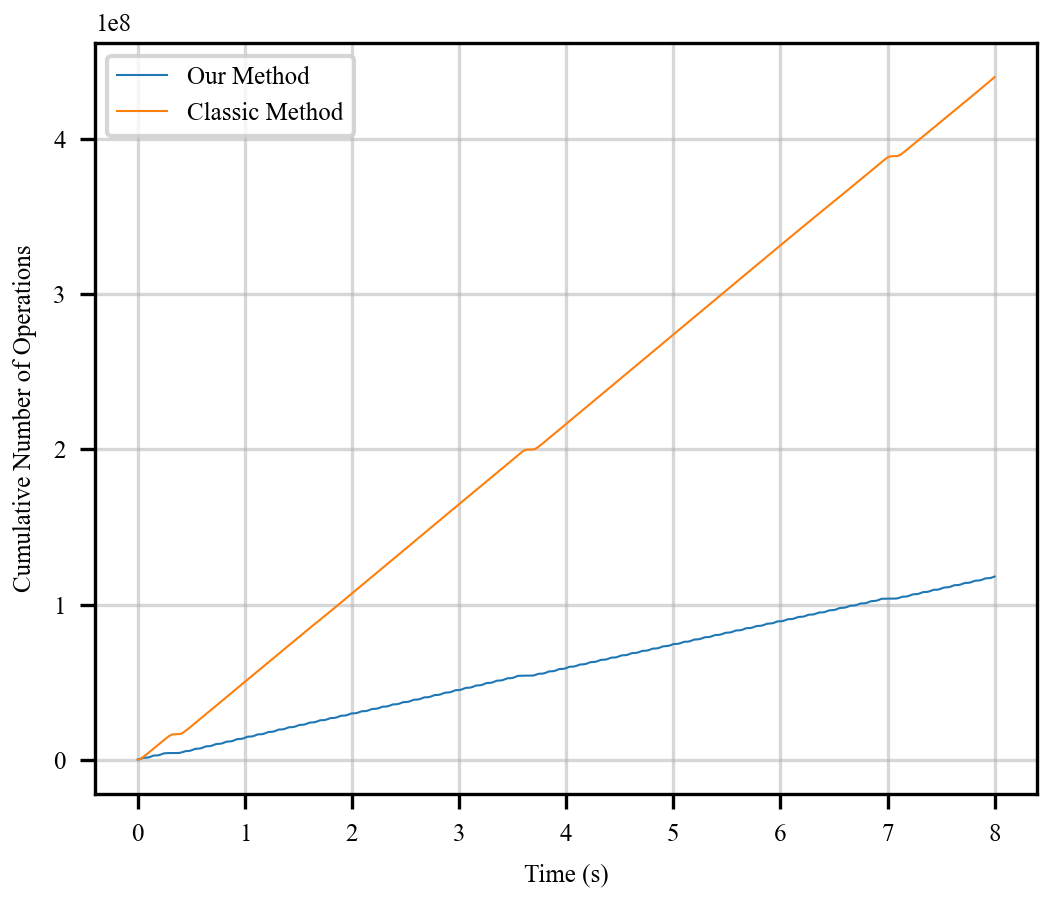}
    \caption{Comparison of cumulative counts of operations on the PandaSet dataset over an $8\ s$ sequence, comparing our method and the conventional method.}
    \label{fig:pandaset_op_count}
\end{figure}

Here we describe the results of our testing on the PandaSet dataset, using the Orin in the same configuration as all earlier tests. First we compare the operation count of the PFN stage. Fig. \ref{fig:pandaset_op_count} shows the corresponding count of operations required to generate the pillar feature maps, comparing SSCATeR and the original PointPillars methods applied to a $8$ second sequence of frames. On average, the SSCATeR PFN stage requires $71.98\%$ fewer operations per frame. It can be seen in Fig. \ref{fig:pandaset_op_count} that the lines plateau occasionally, this is due to sudden absences of data in the raw PandaSet file, for example, between the timestamps $1557540043.853656$ and $1557540043.9501982$, which are also the last and first timestamps respectively of the second and third dataframes in the raw LiDAR data.

\begin{figure}[!h]
    \includegraphics[width=\columnwidth]{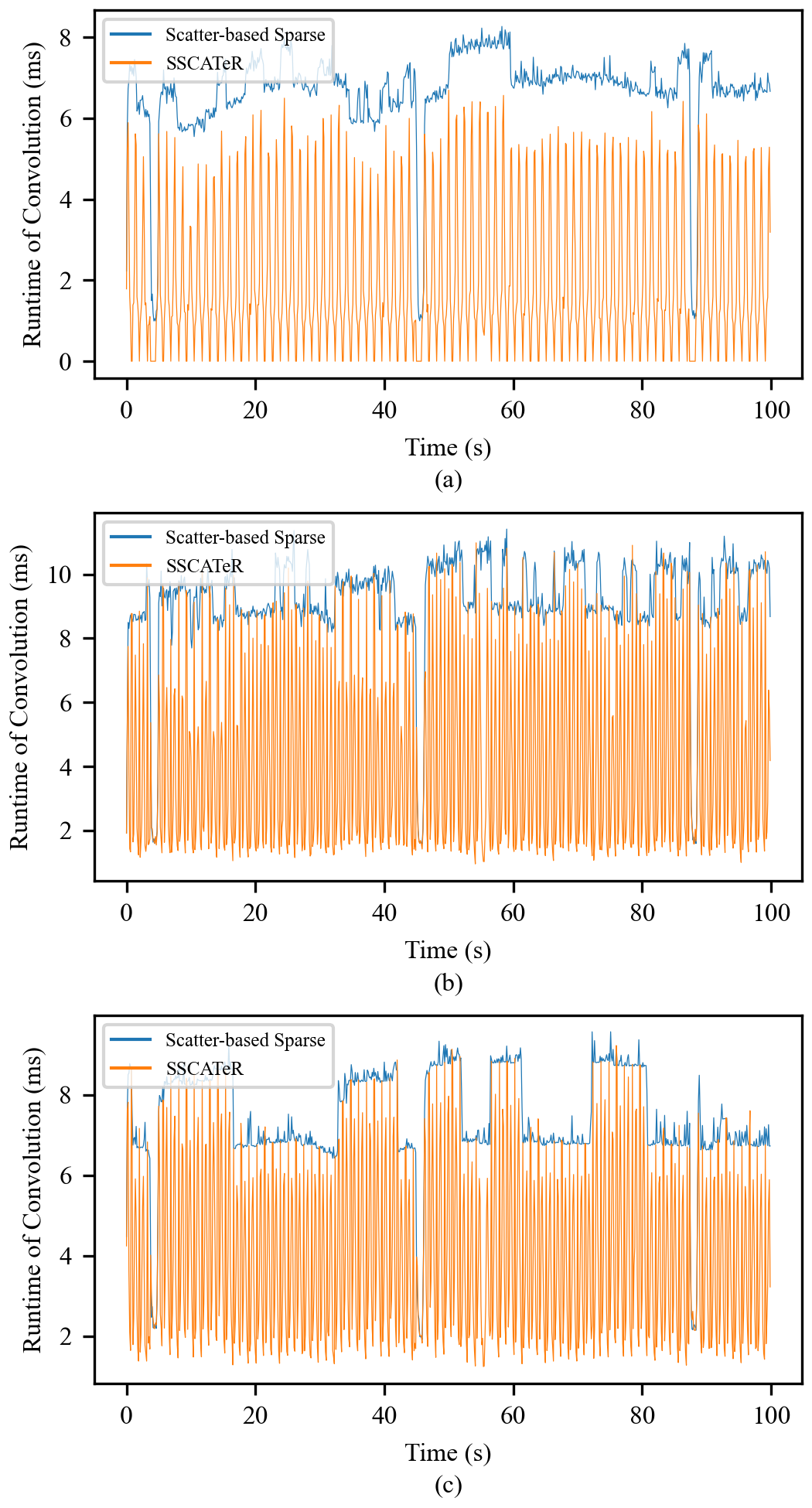}
\caption{Average run time (milliseconds) of SSCATeR on PandaSet dataset sequence compared with scatter-based sparse for each convolutional block (a, b, c) in the backbone on AGX Orin.}
\label{fig:pandaset_backbone}
\end{figure}

In the PandaSet testing, we focus on the full backbone. Fig. \ref{fig:pandaset_backbone} shows the results of each block, and it can be seen that the runtimes are slower than with the drone dataset, although SSCATeR remains consistently faster than a scatter-based sparse backbone. The mean processing time for SSCATeR and the scatter-based sparse backbones were $10.11760\ ms$ and $23.26825\ ms$ respectively. Whilst this means that SSCATeR was slightly slower than real-time on the PandaSet testing, it was still $2.30$ times faster than scatter-based sparse convolution.

\subsection{LiDAR-to-LiDAR Crosstalk Interference Results} \label{sec:crosstalk_results}

\begin{table}[]
\centering
\caption{Comparison of the Impact of Crosstalk on PillarNet and Aerial-PointPillars}
\label{tab:crosstalk_performance}
\begin{tabular}{|p{40pt}|p{42pt}|p{28pt}|p{22pt}|p{25pt}|p{22pt}|}
\hline
Architecture                             & Crosstalk Type        & Precision   & Recall   & F1-Score   & mAP      \\
\hline
\multirow{4}{40pt}{PillarNet\\($0.16\ m$)} & Baseline              & $94.192$    & $94.972$ & $94.580$   & $93.855$ \\
                                         & \textit{No Indirect}  & $94.866$    & $94.972$ & $94.920$   & $93.877$ \\
                                         & \textit{No Direct}    & $94.554$    & $94.972$ & $94.763$   & $93.870$ \\
                                         & \textit{No Crosstalk} & $95.235$    & $94.972$ & $95.103$   & $93.892$ \\
                                         \hline
\multirow{4}{40pt}{Aerial-PointPillars}  & Baseline              & $83.588$    & $93.702$ & $88.356$   & $93.021$ \\
                                         & \textit{No Indirect}  & $84.504$    & $93.702$ & $88.866$   & $93.051$ \\
                                         & \textit{No Direct}    & $84.085$    & $93.702$ & $88.633$   & $93.029$ \\
                                         & \textit{No Crosstalk} & $85.013$    & $93.702$ & $89.146$   & $93.059$ \\
                                         \hline
\end{tabular}
\end{table}

\begin{figure}[bth]
    \includegraphics[width=3.49in]{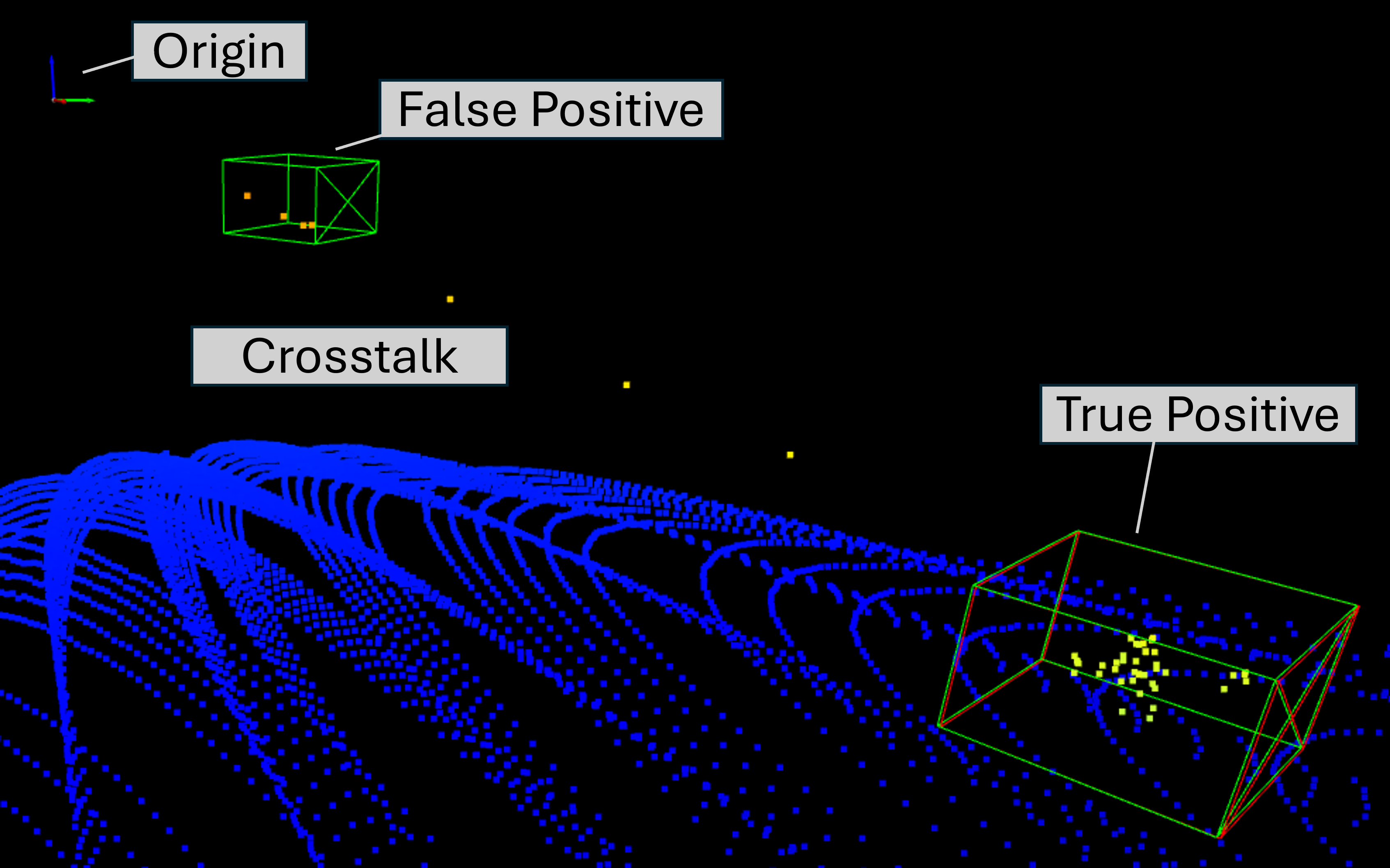}
    \caption{Example of a false positive caused by indirect LiDAR-to-LiDAR crosstalk interference, detected using the trained Aerial-PointPillars model.}
    \label{fig:crosstalk_example}
\end{figure}

This section discusses and analyzes the results of the LiDAR-to-LiDAR crosstalk interference testing, as described in section \ref{sec:crosstalk}. We measure the performance of the PillarNet and Aerial-PointPillars models across a baseline, no direct crosstalk interference, no indirect crosstalk interference, and no crosstalk interference of either type. To categorize false positive predictions as attributable to crosstalk, visual analysis of prediction boxes was conducted, and those found to be containing or directly adjacent to crosstalk points were classified as such. Our testing found no false negatives that could be reasonably attributed to crosstalk. The results of this testing can be found in Table \ref{tab:crosstalk_performance}. For PillarNet, of the $106$ false positives, $20$ were attributed to crosstalk, with seven being caused by direct interference, and twelve from indirect interference. The mAP of the network increases by $0.037$ points, from $93.855\%$ to $93.892\%$. For Aerial-PointPillars, of the $332$ false positives, $33$ were attributed to crosstalk, with twelve being caused by direct interference, and $21$ from indirect interference. The mAP of the network increases by $0.038$ points, from $93.021\%$ to $93.059\%$. An example of a false positive created by crosstalk can be seen in Fig. \ref{fig:crosstalk_example}.

Of the $541$ test frames collected whilst the LiDAR's were directly facing each other, false positives occurred in $1.29\%$ of inference passes for PillarNet and $2.21\%$ for Aerial-PointPillars. Of the $1,269$ frames collected during scenarios where indirect crosstalk was likely to occur, false positives occurred in $1.02\%$ of inference passes for PillarNet and $1.65\%$ for Aerial-PointPillars. Analysis of the point clouds shows that, while direct crosstalk did create much more interference, the intensity of the points was much higher than that of the drone points the network has been trained on, whereas indirect crosstalk intensity was similar to those belonging to a drone, and we believe this to be the cause of the similarity in the occurrence of false positives.

\subsection{Performance and Results Evaluation} \label{sec:evalutation}

SSCATeR is consistently more efficient than scatter-based sparse convolution techniques, scaling well as computational demand increases, whilst producing identical outputs to its sparse or dense counterparts. Therefore, there is potential for SSCATeR to exploit the performance gains of resource intensive networks while using resource-restrictive hardware.

Our results show that a SSCATeR-based backbone performs strongly on the Orin, which fits within the size, weight and power requirements for mounting on a commercial drone, such as a DJI M300, making our network suitable for real-time onboard detection. As is also shown by the highly optimized implementations of CUDA-PointPillars \cite{cuda-pp} and Intel's PointPillars \cite{wang_yang_optimization_2021}, the backbone dominates the runtime of this network. The timing impacts of the other subsystems are negligible in comparison, and thus a SSCATeR-based backbone is preferable for real-time systems.

With this in mind, timings are hardware dependent, and this work shows the need for the weight and power draw increase of the Orin over the Xavier for this task. Our work shows that SSCATeR scales comparably across both options, and its performance enhancements are not limited to just lightweight hardware.

When analyzing the results of the PillarNet testing, using spconv's benchmarking features, it was found that upwards of $40\%$ of the processing time in the sparse convolutions was caused by the reconstructing of the rulebook when the feature map is downsampled. However, it is also seen that even the rulebook-based PillarNet has competitive latency to a scatter-based Aerial-PointPillars, therefore PillarNet could see further runtime improvements at no cost to its impressive performance if SSCATeR is applied to it.

\subsection{Analysis and Discussion of Method Limitations} \label{sec:limitations}

\begin{figure}[bth]
    \includegraphics[width=3.49in]{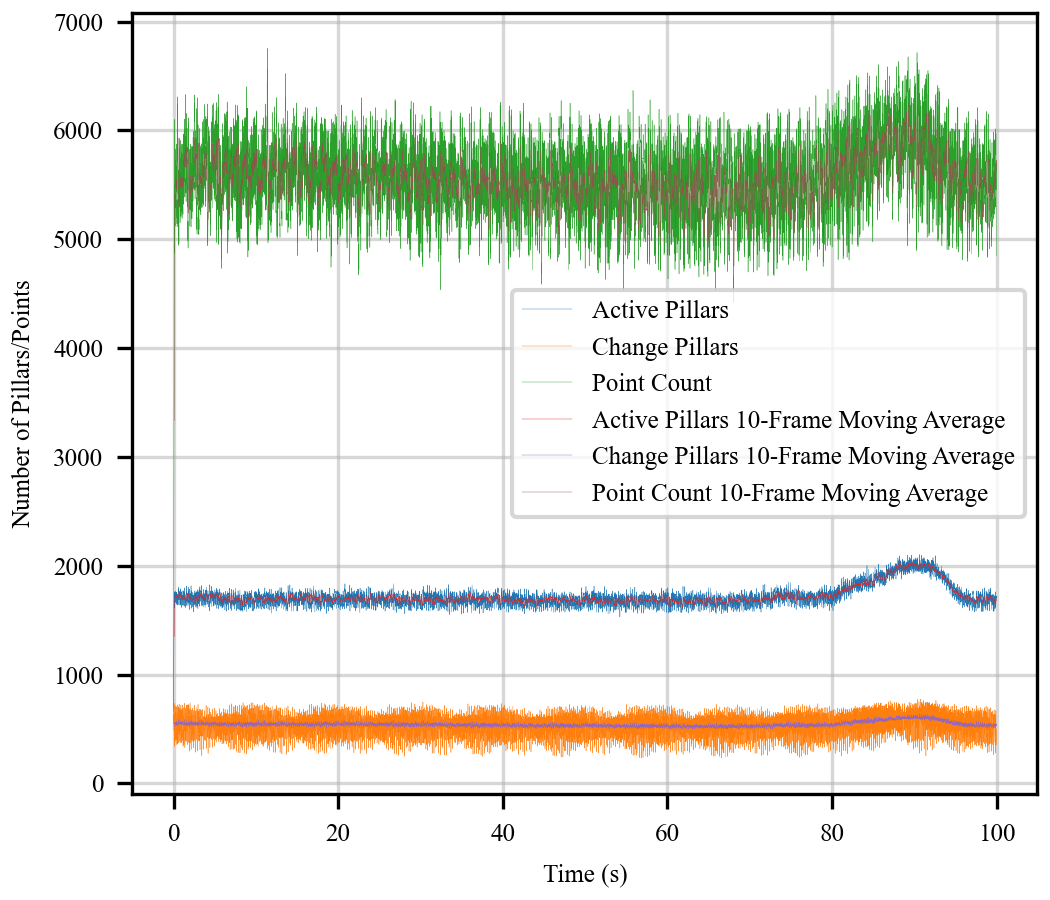}
    \caption{Graph showing the points, active, and changed pillars over time on the Hamburg dataset, which can be seen to correspond to the runtime of the network on the embedded boards.}
    \label{fig:hb_pill_pts_over_time}
\end{figure}

\begin{figure}[bth]
    \includegraphics[width=3.49in]{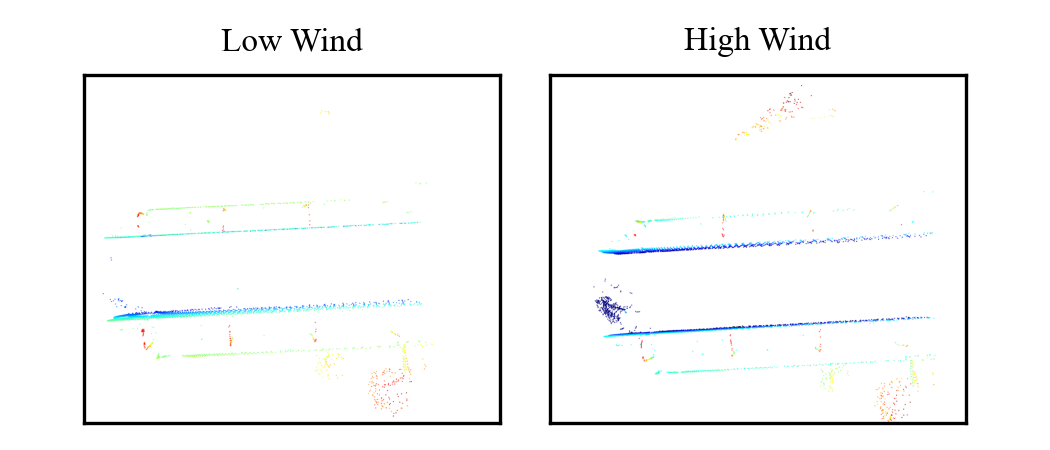}
    \caption{A comparison of point clouds taken from the same position in the Harburg Lock, under low and high wind conditions respectively.}
    \label{fig:lowvhighwind}
\end{figure}

We also analyzed the causes for the increase in runtime on the drone datasets, especially in the mostly stable Hamburg dataset. We note that and increase in incoming points leads to an increase in pillars as shown in Fig. \ref{fig:hb_pill_pts_over_time}. When visualizing the point cloud, it can be seen that due to an increase in wind, the number of points reflected by the water in front of the drone increases drastically, as seen in Fig. \ref{fig:lowvhighwind}. The effects of atmospheric, vibrational, and sudden velocity changes on LiDAR accuracy, such as those caused by wind gusts, are well documented~\cite{pos_error}, ~\cite{wind_blows}, ~\cite{fast_moving}. The stability in runtime seen in the Hamburg dataset compared to the Newcastle makes sense then, as in Hamburg the drone was operating within the lock walls, and better protected from the wind.

\begin{figure}[thb]
    \includegraphics[width=\columnwidth]{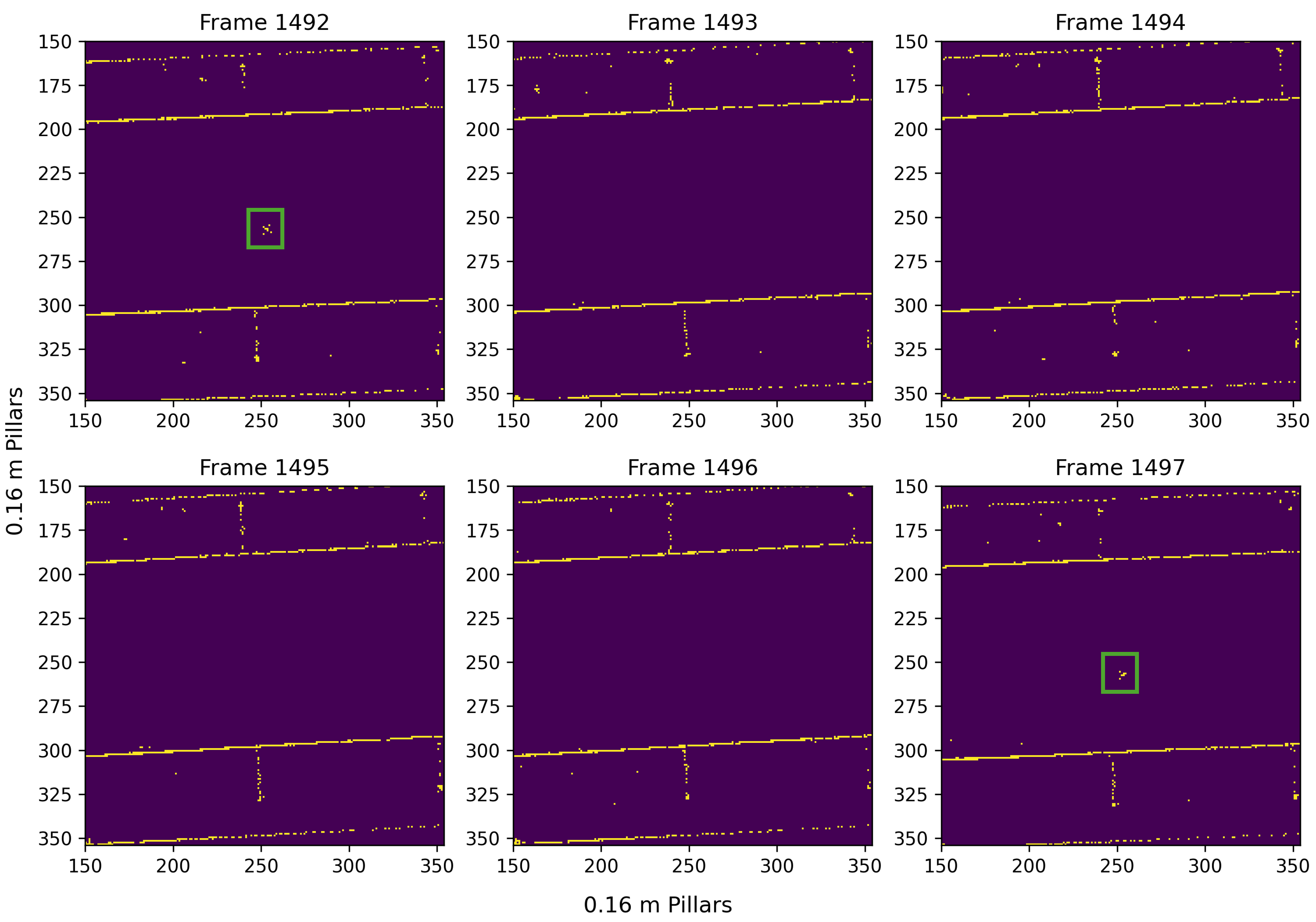}
    \caption{Example of a scenario where the SSCATeR algorithm fails to identify a drone for several frames. The drone, enclosed in a green box in the image, is not visible from frames $1,493$ to $1,496$ as the non-repetitive scanning pattern has not captured any of its points.}
    \label{fig:failure_case}
\end{figure}

There are failure cases for the SSCATeR algorithm, for example, when the LiDAR scanning pattern does not capture points belonging to the drone due to its non-repetitive nature. As shown in Fig. \ref{fig:failure_case}, a missed detection can propagate across multiple frames. Additionally, if too few points are collected in the leading $10\ ms$ window to make a detection, it may be required that subsequent $10\ ms$ frames are required to build up the feature map delaying the time of achieving successful detection. The use of a tracking algorithm could provide a fallback sub-system in the case of a full SAD system. From a latency perspective, SSCATeR performs best when point clouds are extremely sparse, denser point clouds can limit its capability to perform in real-time, and therefore much consideration is required in hardware design, not just in terms of SWaP restrictions and compute power, but additionally for parameters such as pulses per second from the LiDAR unit itself.

As for the increase in runtime on the PandaSet dataset, a considerable amount of points in the scene are returned by the road surface, a scenario that is mostly outside the scope of this method's use case. As such, when points below $10\ cm$ are removed, which accounts for most, but not all, of the road surface, the average points across the sequence reduce from $61,019.23$ to $37,264.05$. With this consideration in mind, SSCATeR has the potential to utilize LiDAR systems which produce more points in an aerial domain, and still perform in real-time on lightweight systems.

Considering the results of the crosstalk testing, we find the interference to be minimally detrimental to object detection performance. Crucially from a safety critical aspect, crosstalk did not affect the false negatives and thus the recall, meaning that the only errors were in the form of additional detections rather than missed detections. That being said, there is still the potential that crosstalk could affect the morphology of a drone's point cloud, and thus impact the recall, as such mission planning with multiple LiDARs must minimize the effects of crosstalk interference. Reducing direct interference is rather straightforward, drones in a swarm should not directly face one another with their LiDAR sensors during operation. The reduction of indirect interference is more complicated and should consider the beamforming of the LiDAR sensor. In our testing, the majority of indirect crosstalk occurred when the center of two sensors were aligned and facing the same direction. Therefore, ensuring that the observer drone is located at a different altitude to the LiDAR drones within the swarm should help reduce the effects of indirect crosstalk. Furthermore, the number of erroneous points caused by the crosstalk interference is limited by the design of the chosen LiDAR units. In Risley prism-based LiDAR units, the emitter and the photodetector used to receive the return sit behind two rotating prisms, meaning that crosstalk interference must enter the prisms at the correct angle to reach the photodetectors. This further supports the use of non-repetitive scanning patterns when using LiDARs in drone swarm-based object detection.

\begin{figure}[thb]
    \includegraphics[width=\columnwidth]{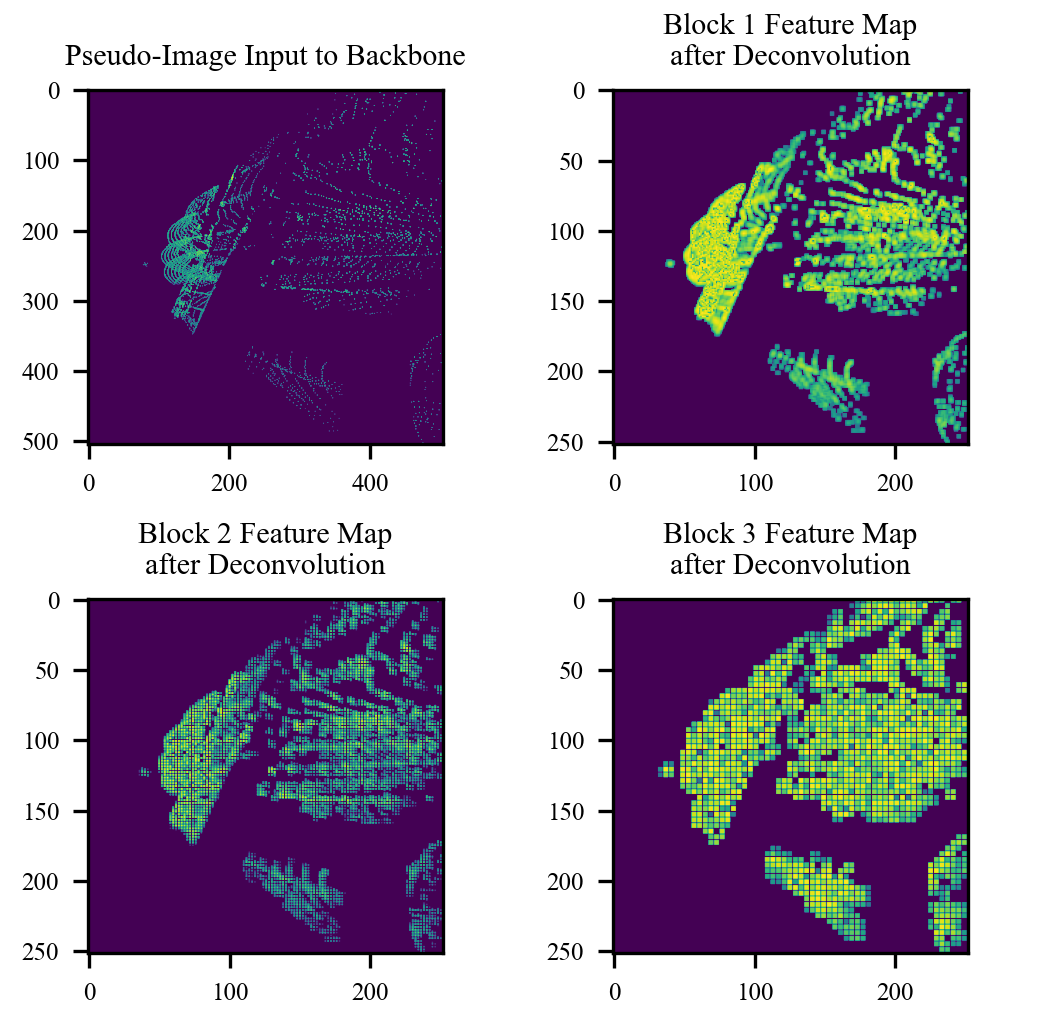}
    \caption{Convolution results of the pseudo-image, block 1, block 2, and block 3, all after deconvolution.}
    \label{fig:conv_visual}
\end{figure}

We display a visual example of convolution results from the Newcastle dataset throughout the network in Fig. \ref{fig:conv_visual}. The input pseudo-image is a $504 \times 504$ canvas of features, and an example of one of the 64 channels output by the 1D convolution process. The three other convolution results come from the convolutional backbone after the upsampling process.

By visualizing the three blocks after transpose convolution, we can see how the difference in granularity can be used to identify features of different sizes. The unique morphology of the drone is best viewed in the finer detail of block 1, whilst in block 3 the larger striding makes it stand out more compared to the other transpose convolution. These results are identical when either rulebook-based sparse convolution, scatter-based sparse convolution or SSCATeR are applied.

\section{Conclusion}
\label{sec:conclusion}

In this work, we have presented a novel data reuse-focused convolution algorithm called SSCATeR. By considering the temporal dimension of LiDAR in combination with a data reuse strategy we have approximately doubled the performance of scatter-based sparse convolution. We tested SSCATeR on two Nvidia embedded compute systems, the AGX Xavier and Orin, with two LiDAR datasets collected by drones, and integrated SSCATeR into the PointPillars object detection network. 

We have demonstrated that SSCATeR generates the same results as both scatter-based and rulebook-based sparse convolution with less memory consumption and eliminates the need for a rulebook in sparse and sparse transpose convolution. Our testing shows that this leads to a mean reduction of $72.80\%$ in the number of sites to be processed per frame. This reduced processing time by as much as $6.61$ times over a single convolutional layer and on average between $1.73$ and $3.88$ times across a whole backbone, with its fastest runtime being $1.731936\ ms$. In the context of a full SAD system, this means the Aerial-PointPillars subsystem would perform in real-time and detect $91.277\%$ of drones, making it a strong contender for integration as a component of a full SAD system.

The timing results achieved by the rulebook-based PillarNet were faster than expected when compared to a rulebook- and scatter-based Aerial-PointPillars.  Future work that applies SSCATeR to other sparse architectures, such as PillarNet, could see further latency reduction whilst retaining the high performance of those networks, and make for interesting potential next steps.

Our GPU implementation follows a gather-GEMM-scatter dataflow \cite{torchsparse}, which needs to read and write DRAM a minimum of three times, respectively. There is significant room for optimization using the following techniques: Firstly, perform computation on on-chip SRAM (shared memory) using tiling, which reduces access to DRAM memory, and secondly, overlap memory access with computation.

\FloatBarrier
\printbibliography

\begin{IEEEbiography}[{\includegraphics[width=1in,height=1.25in,clip,keepaspectratio]{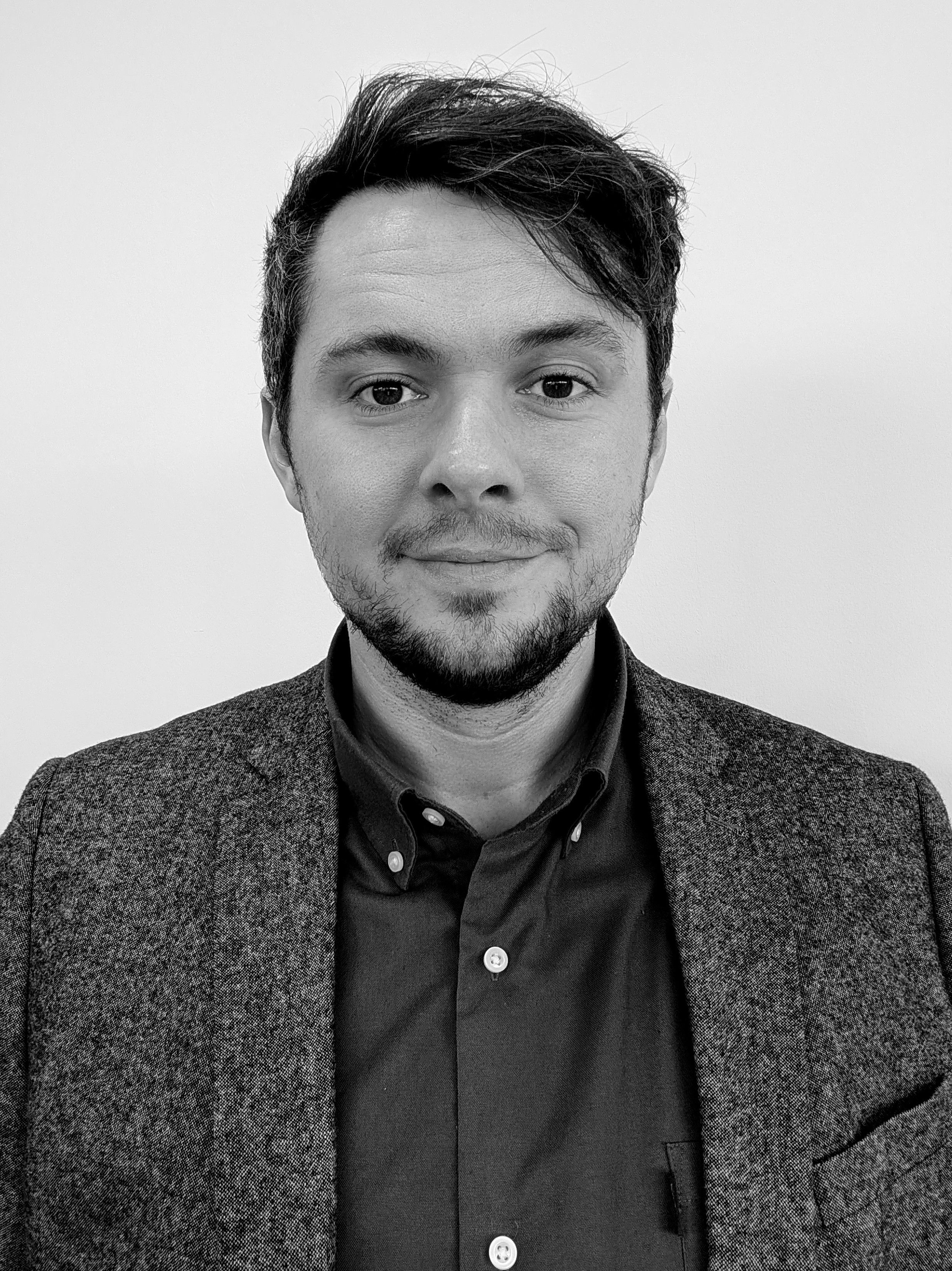}}]{Alexander Dow} 
received his B.S. in sound engineering and production from Birmingham City University,
Birmingham, UK in 2016 and his M.S. in digital signal and image processing from the 
University of Sussex, Brighton, UK in 2020. He recieved his Ph.D. in drone-based computer vision at the University of the West of Scotland, Lanarkshire, Scotland, in 2025.

He currently works as a post-doctoral research assistant in the Drone Systems Lab at the University of the West of Scotland. His research interests include computer vision, artificial intelligence on embedded systems, LiDAR and hyperspectral imaging.

\end{IEEEbiography}

\begin{IEEEbiography}[{\includegraphics[width=1in,height=1.25in,clip,keepaspectratio]{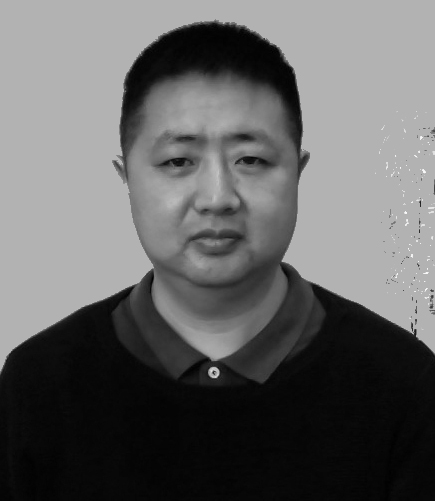}}]{Manduhu Manduhu} received the M.Sc. and
Ph.D. degrees in information engineering from
Hiroshima University, Japan, in 2010 and 2013,
respectively. 

He is currently a post-doctoral researcher at the Drone Systems Lab of the University of the West of Scotland. His research interests include parallel algorithm, physics simulation and 3D object detection.

\end{IEEEbiography}

\begin{IEEEbiography}[{\includegraphics[width=1in,height=1.25in,clip,keepaspectratio]{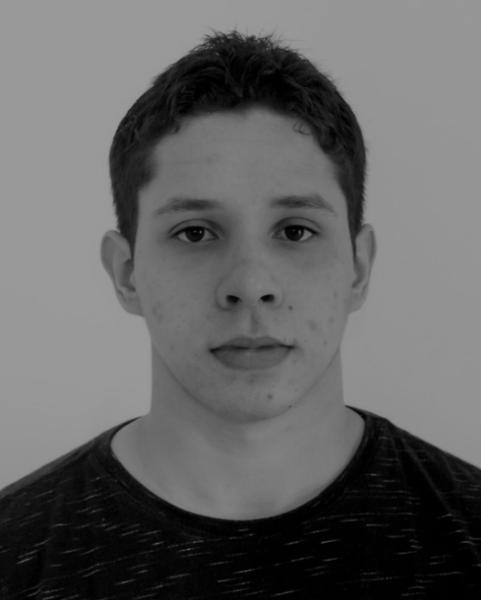}}]{Matheus Santos}
received his B.S. and M.S. degrees in Electronic and Computer Engineering from the Federal University of Sergipe, Brazil. He completed the Ph.D. degree in Electronic and Computer Engineering at the University of Limerick, Ireland, in 2024.

He is currently an Associate Professor B at the Electronic and Computer Engineering Department, University of Limerick, and a member of the Centre for Robotics and Intelligent Systems (CRIS). He is also affiliated with LICHPU, a joint transnational education initiative between the University of Limerick (UL) and Henan Polytechnic University (HPU), China. His research interests include automation, computer vision, control systems, and machine learning, particularly as applied to robotic and autonomous platforms.
\end{IEEEbiography}

\begin{IEEEbiography}[{\includegraphics[width=1in,height=1.25in,clip,keepaspectratio]{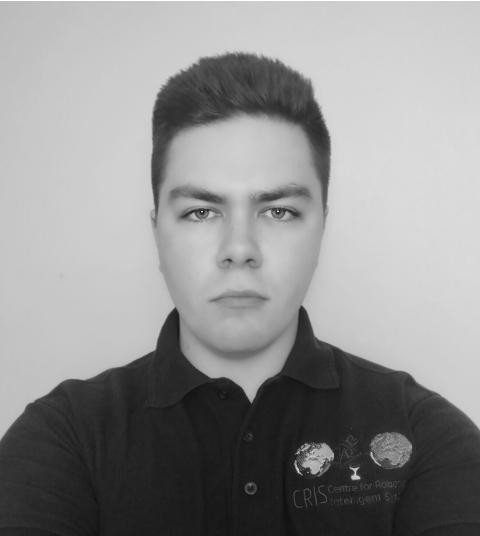}}]{Ben Barlett}
received the B.Eng. degree with first-class honours in Electronic and Computer Engineering from the University of Limerick, Ireland, in 2021, and the Ph.D. degree in 2025 from the Centre for Robotics and Intelligent Systems (CRIS), University of Limerick.

He is currently an Associate Professor B in the Electronic and Computer Engineering Department at the University of Limerick and is affiliated with Limerick International College, Henan Polytechnic University (LICHPU), China. His research interests focus on automation of data acquisition and analysis workflows, machine vision, and the design of heterogeneous sensing systems for large-scale inspection and environmental assessment.

Prof. Bartlett works on building and deploying end-to-end automated systems, with emphasis on robustness, integration, and real-world validation. 
\end{IEEEbiography}

\begin{IEEEbiography}[{\includegraphics[width=1in,height=1.25in,clip,keepaspectratio]{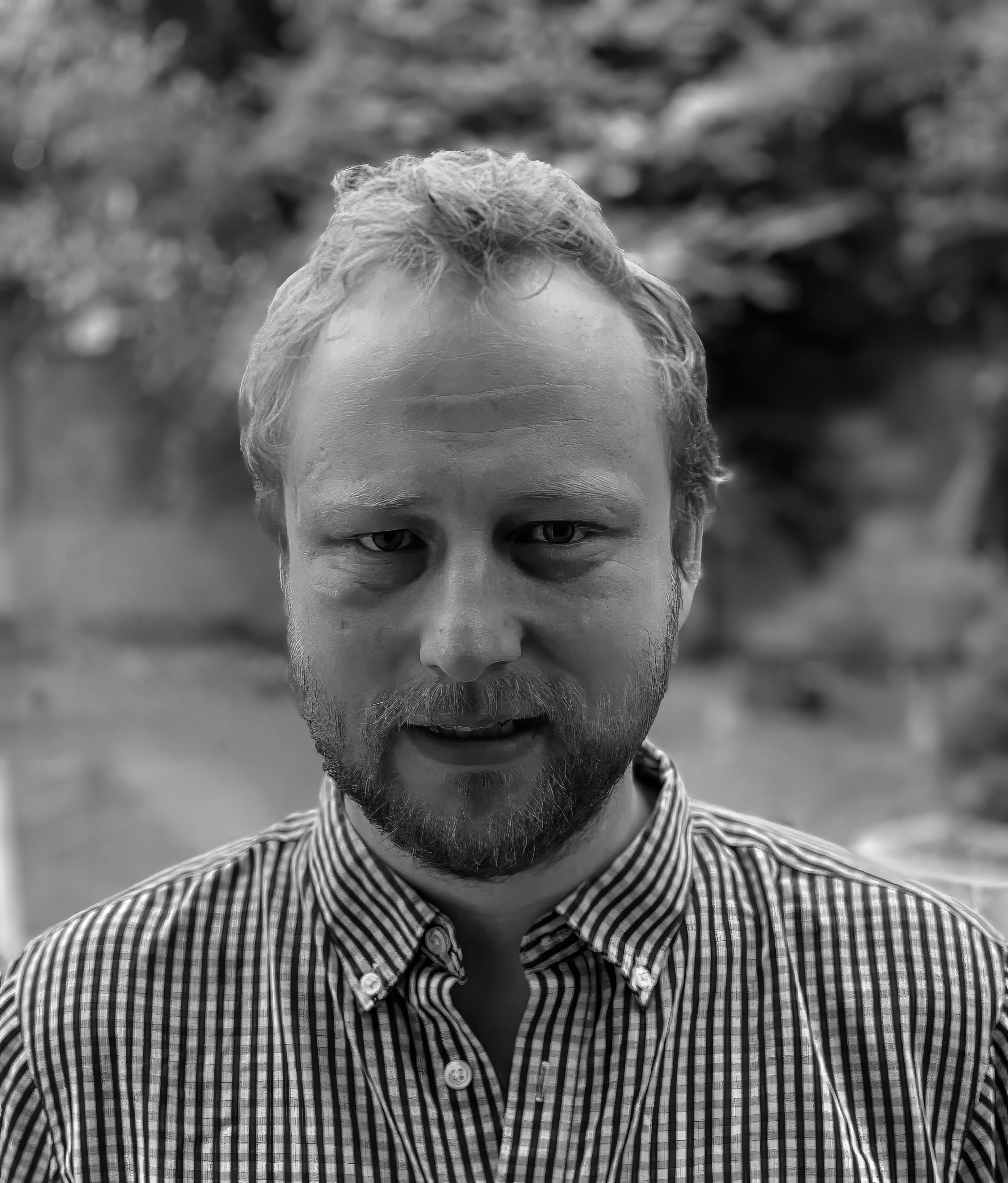}}]{Gerard Dooly}
received the B.Eng. degree from the Electronic and Computer Engineering Department, University of Limerick, in 2003, and the Ph.D. degree from the Optical Fibre Sensors Research Centre, University of Limerick, in 2008, on the topic “An Optical Fibre Sensor for the Measurement of Hazardous Emissions from Land Transport Vehicles”. 

For over ten years, he has worked extensively in field robotics and perception. His research interests include SLAM, machine vision, machine learning, optical fibre sensors, subsea structural health monitoring and teleoperation. He is involved in robotics for harsh environments in offshore settings including development of systems to address beyond visual line of sight operations for UAS. 

Prof. Dooly is focused on the design and development of robotics to address real-world issues and has engaged in numerous platform and project demonstrations both in Ireland and on the continent.
\end{IEEEbiography}

\begin{IEEEbiography}[{\includegraphics[width=1in,height=1.25in,clip,keepaspectratio]{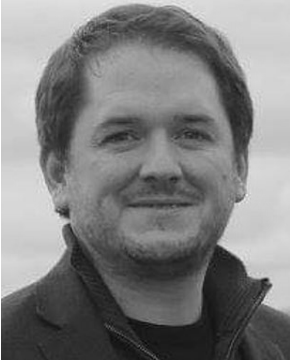}}]{James Riordan}
received the B.Eng. degree in electronic engineering specializing in aircraft simulation
systems and the Ph.D. degree in real-time processing of acoustic signals from the University of
Limerick, Ireland. 

Currently, he is a Full Professor with the University of the West of Scotland, U.K.,
where he is also the Director of the Drone Systems Laboratory. He is also a Principal Investigator
of multiple research projects funded by European
Commission and U.K. Research and Innovation. His research interests include artificial intelligence,
computer vision, and sensing methods to extend the safe and sustainable
application of autonomous vehicles in land, air, and sea.

\end{IEEEbiography}

\end{document}